%% file: sample-base.tex
\documentclass[manuscript]{acmart}
\usepackage{enumitem}
\usepackage{savesym}
\usepackage{graphicx}
\usepackage{CJKutf8}
\usepackage{ulem}
\AtBeginDocument{%
  \providecommand\BibTeX{{%
    \normalfont B\kern-0.5em{\scshape i\kern-0.25em b}\kern-0.8em\TeX}}}

\setcopyright{acmcopyright}
\copyrightyear{2018}
\acmYear{2018}
\acmDOI{XXXXXXX.XXXXXXX}

\acmConference[Conference acronym 'XX]{Make sure to enter the correct
  conference title from your rights confirmation emai}{June 03--05,
  2018}{Woodstock, NY}
\acmPrice{15.00}
\acmISBN{978-1-4503-XXXX-X/18/06}

\acmSubmissionID{5914}

\begin{document}

%%
%% The "title" command has an optional parameter,
%% allowing the author to define a "short title" to be used in page headers.
% \title{TypeDance: Creating Semantic Typographic Logos from Images through Personalized Generation}

\title[TypeDance]{TypeDance: Creating Semantic Typographic Logos from Image through Personalized Generation}

%%
%% The "author" command and its associated commands are used to define
%% the authors and their affiliations.
%% Of note is the shared affiliation of the first two authors, and the
%% "authornote" and "authornotemark" commands
%% used to denote shared contribution to the research.
\author{Shishi Xiao}
% \authornote{Both authors contributed equally to this research.}
\email{sxiao713@connect.hkust-gz.edu.cn}
\orcid{0009-0008-0262-5289}
\affiliation{%
  \institution{The Hong Kong University of Science and Technology (Guangzhou)}
  % \streetaddress{P.O. Box 1212}
  % \city{Dublin}
  % \state{Ohio}
  \country{China}
  % \postcode{43017-6221}
}

\author{Liangwei Wang}
% \authornotemark[1]
\orcid{0000-0003-3481-3993}
\email{lwang344@connect.hkust-gz.edu.cn}
\affiliation{%
  \institution{The Hong Kong University of Science and Technology (Guangzhou)}
  % \streetaddress{P.O. Box 1212}
  % \city{Dublin}
  % \state{Ohio}
  \country{China}
  % \postcode{43017-6221}
}

\author{Xiaojuan Ma}
% \authornotemark[1]
\orcid{0000-0002-9847-7784}
\email{mxj@cse.ust.hk}
\affiliation{%
  \institution{The Hong Kong University of Science and Technology}
  % \streetaddress{P.O. Box 1212}
  % \city{Dublin}
  % \state{Ohio}
  \country{China}
  % \postcode{43017-6221}
}

\author{Wei Zeng}
% \authornotemark[1]
\orcid{0000-0002-5600-8824}
\email{weizeng@ust.hk}
\affiliation{%
  \institution{The Hong Kong University of Science and Technology (Guangzhou)}
  % \streetaddress{P.O. Box 1212}
  % \city{Dublin}
  % \state{Ohio}
  \country{China}
  % \postcode{43017-6221}
}
% Computational Media and Arts Thrust The Hong Kong University of Scienceand Technology (Guangzhou)Guangzhou, Chinasfu663@connect.hkust-gz.edu.cn
%%
%% By default, the full list of authors will be used in the page
%% headers. Often, this list is too long, and will overlap
%% other information printed in the page headers. This command allows
%% the author to define a more concise list
%% of authors' names for this purpose.
\renewcommand{\shortauthors}{Trovato and Tobin, et al.}

%%
%% The abstract is a short summary of the work to be presented in the
%% article.
\begin{abstract}
Semantic typographic logos harmoniously blend typeface and imagery to represent semantic concepts while maintaining legibility. 
% visually
Conventional methods using spatial composition and shape substitution are hindered by the conflicting requirement for achieving seamless spatial fusion between geometrically dissimilar typefaces and semantics.
While recent advances made AI generation of semantic typography possible, the end-to-end approaches exclude designer involvement and disregard personalized design. This paper presents TypeDance, an AI-assisted tool incorporating design rationales with the generative model for personalized semantic typographic logo design. It leverages combinable design priors extracted from uploaded image exemplars and supports type-imagery mapping at various structural granularity, achieving diverse aesthetic designs with flexible control. Additionally, we instantiate a comprehensive design workflow in TypeDance, including ideation, selection, generation, evaluation, and iteration. A two-task user evaluation, including imitation and creation, confirmed the usability of TypeDance in design across different usage scenarios.
\end{abstract}

%%
%% The code below is generated by the tool at http://dl.acm.org/ccs.cfm.
%% Please copy and paste the code instead of the example below.
%%

\begin{CCSXML}
<ccs2012>
   <concept>
       <concept_id>10003120.10003121.10003129</concept_id>
       <concept_desc>Human-centered computing~Interactive systems and tools</concept_desc>
       <concept_significance>500</concept_significance>
       </concept>
   <concept>
       <concept_id>10010405.10010469</concept_id>
       <concept_desc>Applied computing~Arts and humanities</concept_desc>
       <concept_significance>500</concept_significance>
       </concept>
   <concept>
       <concept_id>10010147.10010178</concept_id>
       <concept_desc>Computing methodologies~Artificial intelligence</concept_desc>
       <concept_significance>500</concept_significance>
       </concept>
 </ccs2012>
\end{CCSXML}

\ccsdesc[500]{Human-centered computing~Interactive systems and tools}
\ccsdesc[500]{Applied computing~Arts and humanities}
\ccsdesc[500]{Computing methodologies~Artificial intelligence}

%%
%% Keywords. The author(s) should pick words that accurately describe
%% the work being presented. Separate the keywords with commas.
\keywords{semantic typography, generative model, personalized design}

%% A "teaser" image appears between the author and affiliation
%% information and the body of the document, and typically spans the
%% page.
\begin{teaserfigure}
  \includegraphics[width=\textwidth]{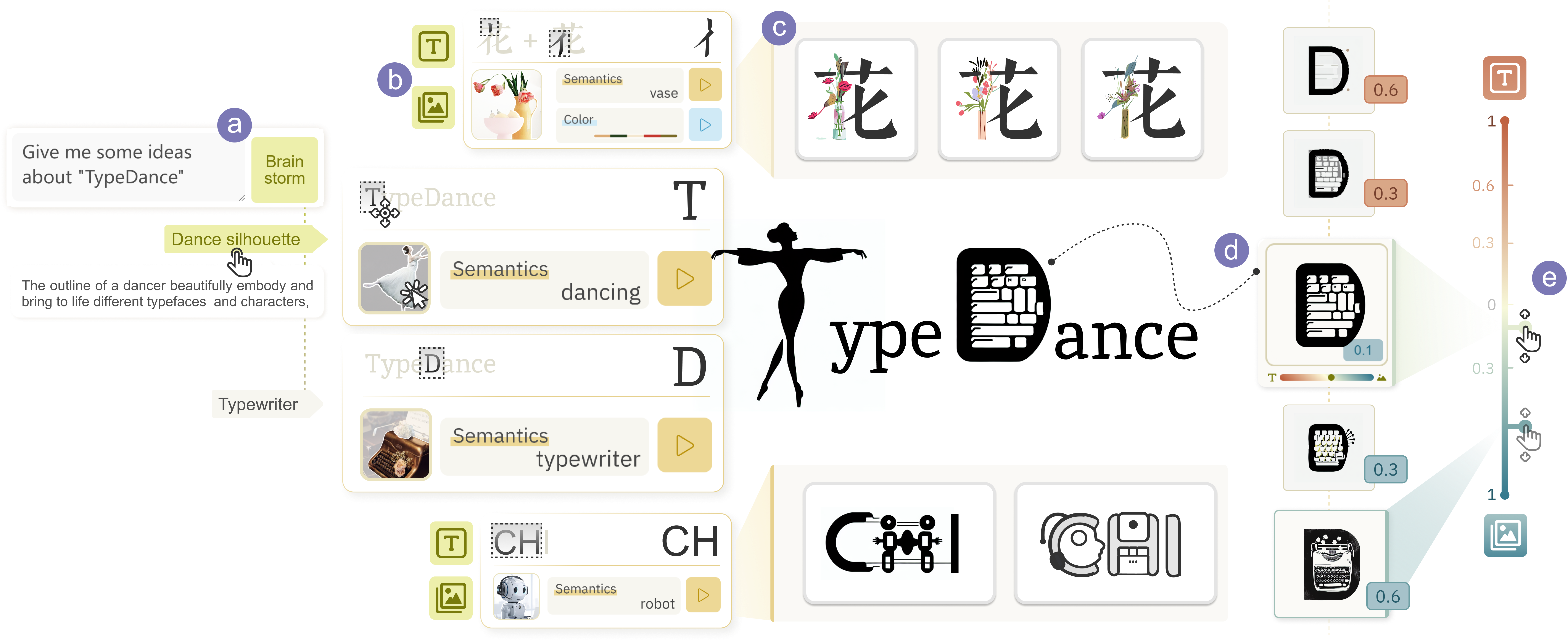}
  \caption{TypeDance is an authoring tool for creating semantic typographic logos with a flexible and personalized design. By distilling the design principles, it instantiates the design workflow and empowers creators to (a) ideate with an interpretable AI agent, (b) select the typeface at different granularity and imagery from images, (c) generate through blending the selected typeface with targeted imagery, then the creator can (d) evaluate and (e) iterate the position of generated result in the type-imagery spectrum.}
  \label{fig:teaser}
\end{teaserfigure}

\received{20 February 2007}
\received[revised]{12 March 2009}
\received[accepted]{5 June 2009}

\newcommand{\eg}{\textit{e.g.}}
\newcommand{\etc}{\textit{etc}}
\newcommand{\ie}{\textit{i.e.}}
\newcommand{\etal}{\textit{et al.}}
\newcommand{\blue}[1]{{\color{black}{#1}}}
\newcommand{\red}[1]{{\color{red}{#1}}}

%%
%% This command processes the author and affiliation and title
%% information and builds the first part of the formatted document.
\maketitle

\input{Latex/1_Introduction}
\input{Latex/2_RelatedWorks}

\input{Latex/3_FormativeStudy}

\input{Latex/4_DesignConsideration}
\input{Latex/5_TypeDance}
\input{Latex/6_Evaluation}
\input{Latex/7_Discussion}
\input{Latex/8_Conclusion}

% \section{Appendices}

%%
%% The next two lines define the bibliography style to be used, and
%% the bibliography file.
\bibliographystyle{ACM-Reference-Format}
\normalem
\bibliography{sample-base}

%%
%% If your work has an appendix, this is the place to put it.
% \appendix

% \section{Research Methods}

% \subsection{Part One}

% Lorem ipsum dolor sit amet, consectetur adipiscing elit. Morbi
% malesuada, quam in pulvinar varius, metus nunc fermentum urna, id
% sollicitudin purus odio sit amet enim. Aliquam ullamcorper eu ipsum
% vel mollis. Curabitur quis dictum nisl. Phasellus vel semper risus, et
% lacinia dolor. Integer ultricies commodo sem nec semper.

% \subsection{Part Two}

% Etiam commodo feugiat nisl pulvinar pellentesque. Etiam auctor sodales
% ligula, non varius nibh pulvinar semper. Suspendisse nec lectus non
% ipsum convallis congue hendrerit vitae sapien. Donec at laoreet
% eros. Vivamus non purus placerat, scelerisque diam eu, cursus
% ante. Etiam aliquam tortor auctor efficitur mattis.

% \section{Online Resources}

% Nam id fermentum dui. Suspendisse sagittis tortor a nulla mollis, in
% pulvinar ex pretium. Sed interdum orci quis metus euismod, et sagittis
% enim maximus. Vestibulum gravida massa ut felis suscipit
% congue. Quisque mattis elit a risus ultrices commodo venenatis eget
% dui. Etiam sagittis eleifend elementum.

% Nam interdum magna at lectus dignissim, ac dignissim lorem
% rhoncus. Maecenas eu arcu ac neque placerat aliquam. Nunc pulvinar
% massa et mattis lacinia.

\end{document}

%% file: Latex/1_Introduction.tex
\section{Introduction}
\label{sec:introduction}

Semantic typography is the art of blending typeface and imagery, where the typeface is conceptualized as a visual illustration of semantic representation with high clarity and legibility~\cite{IluzVinker2023, tanveer2023ds, hermanto2023semantic}.
One notable application is the semantic typographic logo, which symbolizes a unique identity in a concise yet informative manner.
Due to its expressiveness and memorability~\cite{childers2002all}, semantic typographic logo has been widely used as visual signatures for individuals~\cite{lee2015social}, brand logos with commercial values~\cite{henderson1998guidelines, dew2022letting}, and symbols for significant events and city promotions~\cite{roy2022physimorphic, ashworth2009beyond}.

However, crafting a semantic typographic logo presents a formidable challenge, requiring seamless blending of typeface and imagery while preserving readability.
% This requirement introduces a conflict, as typefaces and imagery that are semantically related often have distinct geometric characteristics.
Experienced designers often rely on professional software like Adobe Illustrator to manually adjust the outline of the typeface to incorporate specific imagery, which is a time-consuming and error-prone process.
They often experiment with different strokes or letters of typeface and various imageries to find a visually appealing and memorable representation, intensifying the lengthy process.
% The heterogeneous visual representation of imagery exceptionally increases the complexity, as designers should experiment with different strokes or letters of typeface and various blending combinations to find a visually appealing and memorable representation.
This requires creative thinking, practical skills, and the ability to persist through continuous trial and error.
In addition, the unique identity of a logo necessitates a high level of customization and personalization in the design process. 

\blue{
There are two main challenges in extensive relevant research: blending technique and intent-aware authoring.}
Existing authoring tools leverage various blending techniques to create other types of designs, \textit{e.g.}, graphical icons.
As shown in Fig.~\ref{fig:corpus},
One typical technique aims to spatially composite existing materials~\cite{zhao2020iconate, cunha2020visual}.
Another technique uses shape substitution to achieve a more spatial merge, but it heavily depends on the shape similarity between the objects being blended~\cite{chilton2019visiblends, chilton2021visifit}.
Although some computational design techniques support incorporating imagery into typefaces, they are constrained by the ability to change specific parts of typefaces~\cite{berio2022strokestyles,IluzVinker2023,tendulkar2019trick}.
Advancements in text-to-image generative models~\cite{Rombach_2022_CVPR,zhang2023adding} have made it possible to generate semantic typography automatically, \blue{but it poses another challenge about intent-aware authoring.
Given the myriad details such as specific visual presentation (e.g., semantics, color, and shape) in a logo design, text prompts may not be able to represent these intents.}

This research aims to gain insights into the design space and workflow involved in creating semantic typographic logos, and then instantiate these design principles to create an AI-assisted tool that facilitates personalized generation.
Through analysis of a curated corpus, a systematic design space was identified, focusing on typeface granularity (\ie, stroke-, letter-, and multi-letter-level) and type-imagery mapping (\ie, one-to-one, one-to-many, and many-to-one mappings).
Additionally, interviews were conducted with three experts to gather insights into the challenges and concerns regarding AI collaboration in the design process.
\blue{The findings highlighted the opportunity to simplify the cumbersome blending process and identify a straightforward, explicit material for effectively communicating design intentions to generative models.}
% The findings highlighted the significance of an interactive authoring tool that gives users control over generative models, adheres to the identified design patterns, and offers features for refining and editing the generated designs.

We propose TypeDance, an authoring tool that empowers \blue{both novices and designers with a robust blending technique to create semantic typographic logos from user-customized images.}
\blue{
Delighted from ``\textit{An Image is worth a thousand words}''~\cite{gal2022textual, sangkloy2022sketch} and ``\textit{Everything you see can be design material}''~\cite{devendorf2013anytype, zhang2020dataquilt}, we allow creators to express their design intentions by highlighting the visual representation in their own images.
Meanwhile, multiple design inspirations are extracted from the image references for personalizing design.}
Additionally, we introduce a novel blending technique based on diffusion models that support blending imagery with typeface \blue{at all levels of granularity.
To guarantee the legibility of both typeface and imagery, we harness a vision-language model to assist creators in pinpointing the position of the generated output within the type-imagery spectrum.
We also enable them to edit and refine the output, \textit{e.g.}, making it resemble the typeface ``\textit{D}'' more or adopting typewriter-like imagery, as illustrated in Fig.~\ref{fig:teaser}.}
% The design and development of TypeDance are guided by the workflow outlined in the formative study.
% This approach ensures creators a seamless and intuitive interface to ideate, select, generate, evaluate, and iterate their designs. 
% We have integrated AI collaboration at key stages of the design process to facilitate user-centric design.
To assess the utility of TypeDance, we conduct the baseline comparison and user study with nine novices and nine designers.
Extensive cases and user feedback have revealed the expressiveness of TypeDance in generating a wide range of diverse semantic typographic logos across different scenarios. In summary, we made the following contributions:
\begin{enumerate}
    \item A \textbf{formative study} that identifies generalizable design patterns and simulatable design workflow. 
    \item \blue{An \textbf{intent-aware input} based on user-personalized image that goes beyond ambiguous text prompt, providing a detailed visual description of the desired logo design for generative AI.}
    \item A \textbf{blending technique} that seamlessly incorporates imagery with \blue{all levels of typeface granularity.}
    \item An \textbf{authoring tool} that integrates a comprehensive workflow, \blue{empowering creators to ideate, select, generate, evaluate, and iterate their designs.}
    % \item A two-task evaluation that demonstrates the usability of TypeDance across different usage scenarios.
\end{enumerate}

%% file: Latex/2_RelatedWorks.tex
\section{Related Work}
\label{sec:related-work}

\subsection{Semantic Typographic Logo Design}

Semantic typographic logos are harmonious integration of typeface and imagery, where the imagery is visually illustrated by typeface~\cite{IluzVinker2023, roy2022physimorphic, tendulkar2019trick}.
Compared with plain wordmark~\cite{wang2022aesthetic, wang2020attribute2font} and pictorial logo~\cite{hsiao2023img2logo, li2017rule}, semantic typographic logo allows a more cohesive approach to encode both word and graphic content and enhance the association between them.
The capacity to embody rich symbolism and expressiveness has led to increasing adoption of semantic typographic logos across various scenarios, such as cultural promotion~\cite{ashworth2009beyond}, commercial brand~\cite{dew2022letting} and personal identity~\cite{lee2015social}.
Extensive research has explored how typefaces can be designed to reinforce semantic meaning at varying levels of granularity.
Some studies subdivide typeface into a series skeletal \textbf{\textit{strokes}} with user-guided~\cite{phan2015flexyfont} and automatic segmentation~\cite{berio2022strokestyles}, and then apply structural stylization to each stroke and junction separately.
In contrast, recent studies~\cite{tendulkar2019trick, yang2017awesome, IluzVinker2023} have shifted their focus from stroke-level stylization to individual \textbf{\textit{letter}} stylization using predefined templates. 
For instance, Tendulkar et al.~\cite{tendulkar2019trick} replaced letters with clipart icons relevant to the imagery and visually resembling the corresponding letter. 
Another approach, as demonstrated by Xu et al.~\cite{xu2007calligraphic}, involves compressing the \textbf{\textit{multi-letter}} and arranging them into a predetermined semantic shape.
This approach has been further enhanced by Zou et al.~\cite{zou2016legible}, who proposed an automatic framework that supports the placement, packing, and deformation of compact calligrams.

While prior research extensively investigated the semantic typographic logo across different typeface design granularities, \blue{two key issues persist: 1) these models are constructed for typefaces with specific granularity, limiting their applicability, and 2)} little is known regarding the mapping relationship between typeface and imagery.
These works typically employ a simple approach where one typeface is paired with one specific imagery.
To explore the design space, we collect a real-world corpus, analyze typeface granularity and type-imagery mapping, and instantiate these design principles in TypeDance.
\blue{Then we propose a unified framework based on diffusion model to support flexible blending between imagery and typefaces at different granularities.}

\subsection{Generative Model for Computational Design}

Computational design has garnered considerable attention in the field of generative techniques.
Recently, there have been advancements in aligning semantic meaning between image and text pairs, making natural language a valuable tool that bridges the gap between humans and creativity~\cite{radford2021learning, li2022blip}.
Numerous studies have exploited such semantic consistency to retrieve relevant images from the corpus using natural language statements, which can be used as design materials to generate new designs~\cite{cui2019text, zhao2020iconate}.
While previous studies relied on retrieving from limited corpus and predefined templates, more recent research has proposed text-to-image diffusion models~\cite{ramesh2022hierarchical, tanveer2023ds} that surpass mainstream GAN models\cite{gal2022stylegan} and autoregressive models\cite{ramesh2021zero}. 
However, this plain text-guided generation relies heavily on well-designed prompts, leading to unstable results devoid of user control.
To address this issue and enhance user customization, recent advancements have introduced image-based conditions for achieving controllable manipulations, including depthmap~\cite{Rombach_2022_CVPR} and edgemap~\cite{zhang2023adding}.
Some generative models focusing on font stylization only support the letter-level generation~\cite {IluzVinker2023} and require collecting images containing the specific imagery for fine-tuning the model~\cite{tanveer2023ds}.

While prior works have demonstrated incredible generative ability in creating complex structures and meaningful semantics, ensuring the readability of both the typeface and the imagery remains a daunting task. 
In particular, the text condition lacks sufficient restrictions to capture all user intentions, while the image condition is overly rigid and cannot accommodate the inclusion of additional information.
To tackle this challenge, Mou et al.~\cite{mou2023t2i} proposed an approach that combines multiple conditions to improve controllability.
Similarly, Vistylist~\cite{shi2022supporting} disentangles the design space, enabling the generation with combined user-intended design factors. 
TypeDance builds upon these previous research efforts by providing several design priors that allude to the characteristics of semantic typographic logos. 
These design priors extracted from user-provided images serve as guidance for users to select and incorporate into their designs.
With support for both text and image conditions, TypeDance empowers users with flexible control, enabling personalized and distinctive design outcomes.

\subsection{Graphic Design Authoring Tool}

Significant works have developed authoring tools to facilitate graphic design, which can be broadly divided into two primary categories: ideation and creation tools.
In the domain of ideation, several research studies~\cite{kang2021metamap, xu2021ideaterelate, koch2019may} have proposed interfaces aimed at inspiring ideas and facilitating the exploration of design materials. 
For example, MetaMap~\cite{kang2021metamap} employed a mindmap-like structure encompassing three design dimensions to stimulate users and encourage them to generate a wide range of unique and varied ideas.
Regarding the creation process, as Xiao et al.~\cite{xiao2023let} identified, mainstream works follow a two-stage pipeline, which involves retrieving examples and adapting them as design material~\cite{zhang2020dataquilt} and style transfer reference~\cite{shi2022supporting}.
More recently, researchers sought to blend approaches to create a novel design based on existing design materials.
During the process, spatially compositing semantically related icons to generate a compound design in a resourceful manner is adopted by some researchers~\cite{zhao2020iconate, cunha2020visual}.
Similarly, Zhang et al.~\cite{zhang2017synthesizing} demonstrated that compositing coherent imagery elements can create an ornamental typeface with wide conceptual coverage.
On the other hand, Chilton et al.~\cite{chilton2019visiblends, chilton2021visifit} further explored the potential of blending through similar shape substitution.
For instance, they showed that the ``Starbucks logo'' can replace the position of the ``sun'' as both have a circular shape.

However, spatial composition and shape substitution techniques encounter challenges when dealing with the complexity of semantic typographic logos, in which typeface and imagery need to be spatially fused as a whole despite the absence of shape similarity.
In this work, Typedance utilizes diffusion models to incorporate imagery detail while preserving the salient representation of the typeface, enabling a more natural blend.
Additionally, Typedance integrates both ideation and creation functions. To ensure the readability of both the typeface and the imagery in semantic typographic logos, an evaluation component is further incorporated, enhancing the faithfulness of the design process.

%% file: Latex/3_FormativeStudy.tex
\section{Formative Study}
\label{sec:formative-study}

To instantiate the real design principles in TypeDance, we extracted \textit{simulatable design workflow} from semi-structured interviews and \textit{generalizable design patterns} from a corpus analysis.

\noindent\textbf{Participants.}
We conducted semi-structured interviews with three experts: 
a design professor \blue{leading logo design teams for international conferences and city identities} (E1), a brand designer with over 11 years of \blue{corporate and startup logo} design experience (E2), and a logo designer \blue{who has received several renowned design prizes} (E3). All three experts have extensive experience in semantic typography design.

% including a design professor working in an art school who leads several logo design teams for international conference and city identity (E1), a brand designer with over 11 years of logo design experience for corporation and startup (E2), and a logo designer who has received several renowned design prizes (E3).
% All three experts possess extensive experience in semantic typography design.

\noindent\textbf{Procedure.}
Each individual interview, lasting one to one and a half hours, began with a presentation of the interviewee's work from social media. We then delved into their interpretation and detailed explanation of the design process. Finally, we posed questions about key steps in creating a semantic typographic logo, the most challenging step, and expectations and concerns regarding generative models.

% Each individual interview lasted from one to one and a half hours. During each interview, we started by showcasing the interviewee's representative work that they had posted on social media. We then asked them to interpret the work and explain its design process in detail.
% Finally, we asked them the following questions: 
% (1) What are the key steps in creating a semantic typographic logo? 
% (2) Which step in the creation process do you find the most challenging, and why? 
% (3) What expectations and concerns would you have if a generative model were involved?

% \noindent\textbf{Findings.}
% During the interview, we discovered typical \textit{design patterns} (Sect. \ref{ssec:corpus-analysis}), which were further supported by a systematic corpus analysis.
% Additionally, we gained insight into the general \textit{design workflow and challenges} (Sect. \ref{sec: workflow and challenge}) and gained an understanding of their primary \textit{expectation and concerns} regarding AI involvement (Sect. \ref{sec: concern}).

% \begin{figure}[h]
%   \centering
%   \includegraphics[width=0.99\linewidth]{figures/research_pipeline.pdf}
%   \caption{\red{An overall research pipeline of Metaphoraction.}}
%   \label{fig:pipline}
% \end{figure}

\subsection{General Workflow and Challenges}
\label{sec: workflow and challenge}

% The interview showed that a general workflow was established in all experts' design practices.
% We summarized the key aspects of this workflow as well as the challenges associated with it.

\subsubsection{General Design Workflow}
\label{ssec: workflow}
\begin{figure}[t]
  \centering
  \includegraphics[width=0.99\linewidth]{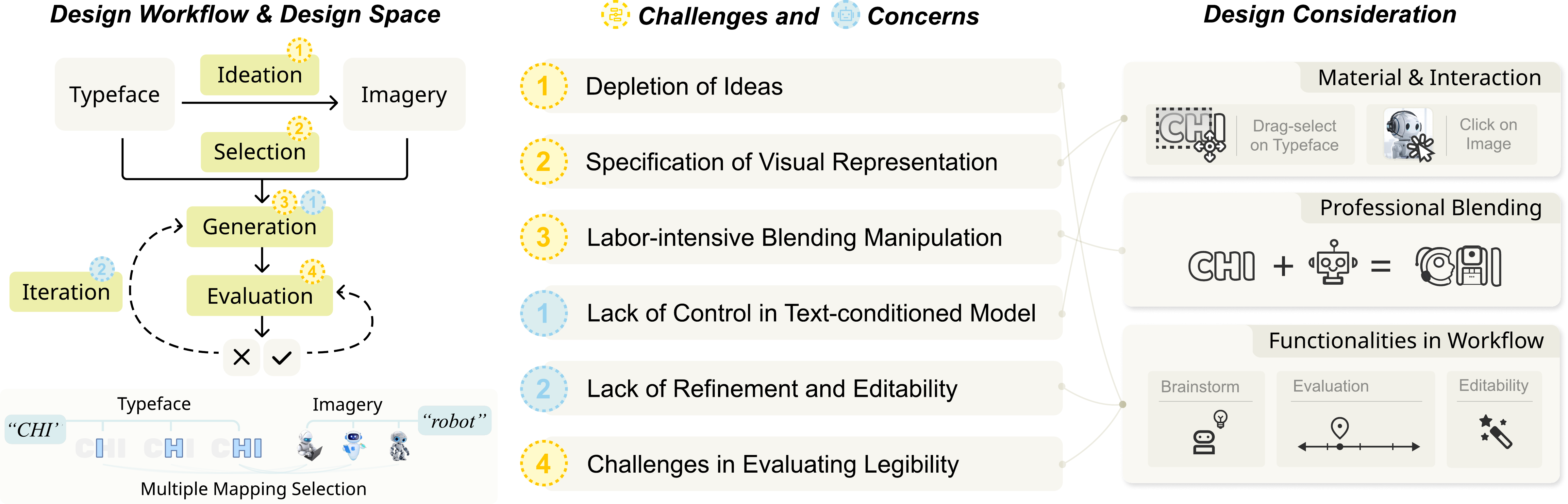}
  \caption{A general workflow for semantic typographic logo design is outlined from the expert interview, with the main challenges in the workflow and concerns of generative AI labeled on corresponding stages. 
Based on the workflow, challenges, and concerns, the design consideration of Typedance is solidified.}
  \label{fig:workflow}
\end{figure}

Semantic typography design is a creative process that involves generating innovative ideas and implementing them. 
As Fig.~\ref{fig:workflow} shows, The general workflow typically consists of five steps.

\begin{itemize}[leftmargin=5.5mm]
\item 
\textbf{Ideation.}
% To begin with, designers are often given fixed text, such as a brand name. Based on the meaning of the text and the usage scenarios, they are required to come up with ``\textit{off-the-wall}'' ideas regarding potential imagery. This process engages in divergent thinking to collect diverse and innovative plans for alternatives.
To begin with, designers are often given fixed text, such as a brand name. Based on the usage scenarios, they need to come up with innovative ideas regarding potential imagery. 
% This process engages in divergent thinking to collect diverse and innovative plans for alternatives.

\item 
\textbf{Selection.}
Once designers confirm a design idea, they prepare design materials with two crucial aspects: 1) the part of the typeface structure and 2) the specific visual representation of imagery. 
\blue{The choice of typeface evolves through multiple attempts with different granularities, such as a single stroke or the entire typeface.
All experts mentioned that they obtained inspiration for visual representation from images, whether provided by customers, collected from design-sharing communities, or from their own life gallery. Images serve as a source of other inspiration as well. As E1 noted, ``\textit{I usually discover new inspiration in pictures, such as color palette.}''}

% Rather than relying solely on imagination, designers utilize search engines and design-sharing communities to gather examples with concrete visual characteristics, which also provide inspiration regarding color and shape.
% The selection of design materials is not fixed but dynamically changes throughout the design process. 

\item
\textbf{Generation.}
% By referring to visual representations in collected exemplars, designers abstract the visual representation into a simple shape that aligns with the skeleton of the typeface.
Designers start by simplifying the visual representation into a basic shape that corresponds to the typeface's skeleton in the sketch. They then adjust the typeface outline using professional software like Adobe Illustrator, seamlessly integrating it with the chosen imagery.
% This process is time-consuming and involves multiple rounds of trial and error.

\item 
\textbf{Evaluation.}
% Design is a highly subjective process influenced by personal preference, especially when it comes to ensuring the legibility of both typeface and imagery in semantic typography.
% To address this subjectivity, designers often seek external valiadation to ensure that the two elements can be understood by people.
Upon completing a design, designers will assess the legibility of both the typeface and imagery incorporated in their work. It often hinges on external validation from individuals other than the designers themselves.

\item 
\textbf{Iteration.}
Iteration is conducted throughout the design process.
Designers conduct multiple experiments and refinements at each step to reach potential outcomes. Refinement continues until a finalized design is achieved.

\end{itemize}

\subsubsection{Challenges in Workflow}
\label{ssec: challenge}

\begin{itemize}[leftmargin=5.5mm,label={\raisebox{-.2\height}{\includegraphics[width=0.4cm]{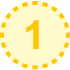}}}]
\item 
\textbf{Depletion of ideas during brainstorming.}
Experts widely regard the birth of good inspiration as a combination of imagination and serendipity.
% This process engages in divergent thinking to collect diverse and innovative plans for alternatives.
E2 emphasizes the importance of delving into the story behind a brand and then incorporating it into logo design.
\blue{This involves gathering background knowledge and discovering imagery that resonates with human perception. Diverse plans for alternatives are essential for iteration.}
% When tackling unfamiliar subjects, extra time needs to be invested in gathering pertinent background knowledge before brainstorming.
% Moreover, considering the imagery's compatibility with the typeface can hinder idea generation.
% During the design process, designers often fall into the trap of ``\textit{design fixation}'' and regurgitate old ideas.
% To overcome this, they must refresh their minds and restart brainstorming to generate new ``\textit{of-the-wall}'' ideas.

\item[\raisebox{-.25\height}{\includegraphics[width=0.4cm]{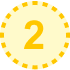}}]
\textbf{Clarification \blue{of specific visual representation of imagery}.} 
Designers frequently explore diverse parts of the typeface while experimenting with various imagery options to attain the most optimal and visually appealing outcomes. \blue{However, a single image can be depicted in various visual presentations, as illustrated by the diverse robots in Fig.~\ref{fig:workflow}, posing a challenge in specifying a particular one. Additionally, the compatibility of the imagery with the typeface can complicate the selection process.}
% For instance, the visual representation of a cat can be depicted either sitting or sprinting.
% Thus, choosing a combination from vast possibilities and validating it as a draft that reflects designers' visions and expectations makes the selection process challenging.

\item[\raisebox{-.25\height}{\includegraphics[width=0.4cm]{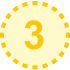}}]
\textbf{Tedious and laborious manipulation in blending typeface and imagery.}
Crafting a logo design from a draft involves \blue{conceptualizing} and professionally \textit{implementing} the blending of significantly different imagery and typeface.
\blue{As mentioned by E3, ``\textit{sometimes it is challenging when my client insists on having both the letter `M' and a standing cat from the photo she gave me – finding similarities in their shapes is hard}.''}
For the implementation, despite relying on professional software like CorelDRAW and Adobe Illustrator, the blending process remains manual, consuming a substantial amount of time and effort.

% During the process, designers are required to manually extract the concise expression of the imagery and align it with the skeleton of the typeface. 
% This involves painstakingly adjusting the outline of vector graphics, a task that consumes significant time and effort. 
% Furthermore, designers may repeatedly experiment with various artistic techniques, such as negative space design, which necessitates a repetitive trial-and-error process.
% has already be specified 

\item[\raisebox{-.25\height}{\includegraphics[width=0.4cm]{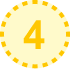}}]
\textbf{Challenges in evaluating legibility.}
A successful semantic typographic logo should not confuse the public about its typeface or meaning. However, evaluating the legibility often falls to designers' subjective judgment rather than being based on the perception of the general public. This poses a challenge in achieving a fair evaluation.

\end{itemize}

\subsection{Concerns in Generative Model Involvement}
\label{sec: concern}

With the rise of generative models like Midjourney, many designers began to embrace AI for design assistance. 
The following are two main concerns about AI involvement in their design process.

\begin{itemize}[leftmargin=5.5mm,label={\raisebox{-.2\height}
{\includegraphics[width=0.4cm]{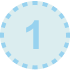}}}]
\item
\textbf{Lack of controllability of text-conditioned generative model.}
The most widely used method for controlling generative models is through textual prompts.
Though various tools can assist in designing prompts, like PromptBase\footnote{\href{https://promptbase.com/}{https://promptbase.com/}} and PromptHero\footnote{\href {https://prompthero.com/}{https://prompthero.com/}}, these tools emphasize imitating certain styles and lacking precise control of specific shape and layout.
When dealing with highly personalized imagery, the time and effort invested in crafting a prompt experiences a significant surge. As E1 notes, \textit{``Each time I intricately design a prompt in anticipation of the generated result, it prompts me to open Illustrator again.}''
% Even taking a lot of time to craft an elaborated prompt, it can not represent 
% While other tools use Controlnet for shape control of the generation, they are typically tailored for character or interior design rather than concise logos.
% Furthermore, this edge-based generation mainly merges the imagery by adding color and texture, not harmonious blending from two shapes.
% As E1 notes, \textit{``I feel like I can't control AI tools. It's not as efficient as the traditional design process.''

\item[\raisebox{-.25\height}{\includegraphics[width=0.4cm]{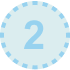}}]
\textbf{Lack of refinement and editability of generative result.}
Lack of editability in the results appears to be a common issue of generated models.
While users may be generally satisfied with the overall outcome, there can be instances where certain details may not meet their expectations (e.g., dislike colors, redundant objects).
One approach to address this is to regenerate the entire image, which results in losing the current design.
\end{itemize}

\subsection{Design Space of Semantic Typography Work}
\label{ssec:corpus-analysis}

\begin{figure}[t]
  \centering
  \includegraphics[width=\linewidth]{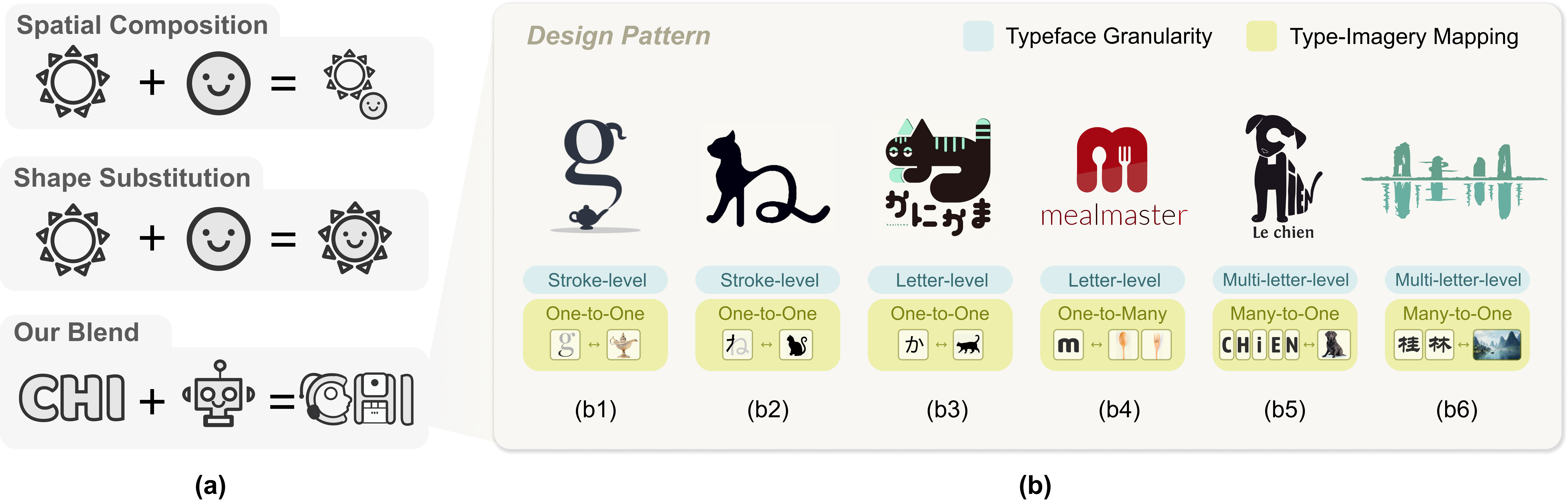}
  \caption{(a) The comparison of blending technique between prior works and TypeDance. (b) Some semantic typographic logo examples from the corpus, each labeled with the corresponding design pattern, including Typeface Granularity and Type-Imagery Mapping.}
  \label{fig:corpus}
\end{figure}

% Through the work interpretation with experts in the interview, we obtained some insights about creating engaging semantic typographic logos. 
To further identify design patterns shaping semantic typographic logos, we collect and analyze a corpus of 427 real-world examples\footnote{\href {https://www.notion.so/Semantic-Typographic-Logo-Dataset-4f552067dbe543ddaa4c630d629502d2}{The online page of a corpus of semantic typographic logo}}.
\blue{To ensure the diversity of sources,} we include examples from prior research~\cite{IluzVinker2023, zhang2017synthesizing}, reputable design communities\footnote{\href{https://www.pinterest.com/}{https://www.pinterest.com/}}, and influential design shared on social media.
% like Pinterest\footnote{\href{https://www.pinterest.com/}{https://www.pinterest.com/}} and Dribbble\footnote{\href{https://dribbble.com/}{https://dribbble.com/}}
The keywords we mainly use for search are \textit{``topographic logo,'' ``semantic topography,''} and \textit{``word as image''}.
\blue{Focusing on logogram (Chinese ($97$), Japanese ($34$), Korea ($30$)) and alphabet language (English ($229$), French($20$), Russian($17$)), we filter search engine results for each language based on ``popular'' and ``new'' criteria, considering both widespread acknowledgment and timeliness.}
As Fig.~\ref{fig:corpus} shows, our corpus analysis reveals two critical aspects of design patterns: (1) typeface granularity and (2) type-imagery mapping.
% We categorized these patterns as \textit{Typeface Granularity} and \textit{Type-Imagery Mapping} and explained them in detail in the following.

\subsubsection{Typeface Granularity}
\label{subsec:typeface-granularity}
While various languages have their own unique symbols, they conform to a shared structural granularity.
% We have identified three levels of granularity that are utilized to convey the single semantic concept.
This hierarchy, arranged from local to global, encompasses \textit{stroke}, \textit{letter}, and \textit{multi-letter}, respectively.

\begin{itemize}[leftmargin=5.5mm]
\item
\textbf{Stroke-level.}
It often employs stroke decomposition, where a single stroke or a group of strokes is associated with imagery ($123/427$ examples).
As illustrated in Fig.~\ref{fig:corpus} (b1), the spout of a teapot aligns with the original curve in the letter ``\textit{g}'', enhancing semantic expression while preserving typeface integrity.
As the smallest unit of typeface, stroke-level blend can be implemented multiple times within a typeface to enrich visual representation.

\item
\textbf{Letter-level.}
Individual letters are commonly used in our collected corpus ($189/427$ examples).
Rather than conducting a letter-level blend to every single letter, certain examples tend to focus on partially representative letters within a word, particularly the first letter in Fig.~\ref{fig:corpus} (b6).
To emphasize the imagery, they employ techniques such as scaling, elongation, and rotation on the letters.

\item
\textbf{Multi-letter-level.}
Blending imagery with multiple letters or entire words is the multi-letter-level blend ($115/427$ examples).
It regards the typeface as a cohesive unit and spatially arranges the letter in a proper position.
According to Fig.~\ref{fig:corpus} (b5), the letters are rearranged and distorted to create the recognizable silhouette of a dog.
% Assigning different parts of a dog to each letter through transformation results in a comprehensive visual representation.
\end{itemize}

\subsubsection{Type-Imagery Mapping}
\label{ssec:type-semantic-mapping}
We observe a complex linkage between typeface and imagery. 
Prior works simplify such linkage by adopting a single mapping strategy, where one letter is associated with specific imagery.
To approximate TypeDance to the design practice of semantic typographic logo, we manually encode the corpus and identify three typical mapping patterns: \textit{one-to-one}, \textit{one-to-many}, \textit{many-to-one}.
With a comprehensive understanding of how typeface interact with imagery, we can better instantiate the design principles to empower the creation process.

\begin{itemize}[leftmargin=5.5mm]
\item
\textbf{One-to-One Mapping.}
One logo corresponds to one imagery commonly observed in the corpus ($294/427$ examples).
It preserves typeface structures using partial strokes or letters to represent the same imagery.
For instance, in Fig.~\ref{fig:corpus} (b1), the letter ``g'' incorporates the imagery of a teapot spout. 
Additionally, we observed that logos with one-to-one mapping often employ repetitive imagery with a consistent style within a particular typeface.

\item
\noindent\textbf{One-to-Many Mapping.}
The semantic typographic logo in this portion will distribute multiple imageries in the typeface involved in the design ($14/427$ examples).
This mapping type supports rich imagery coverage within a compact space, where the semantic concepts usually share the same theme.
Fig.~\ref{fig:corpus} (b4) integrates both spoon and fork into a letter ``\textit{m}'', underscoring the theme of the meal.

\item
\noindent\textbf{Many-to-One Mapping.}
Another aspect involves integrating multiple letters in typeface into a single imagery ($119/427$ examples), typically achieved by combining entire words to convey a complex visual representation of meaning, see Fig.~\ref{fig:corpus} (b3, b4).
This creative approach can be traced back to Giuseppe Arcimboldo, who skillfully merged various elements and shapes to create cohesive portraits and figures~\cite{italypaiter}.
The many-to-one mapping enhances the overall unity and deepens the expression of semantic meaning.
\end{itemize}

\subsubsection{Summary}
\label{ssec:corpus-summary}
Through corpus analysis, we uncover different typeface granularity and various type-imagery mapping.
The combination of these design patterns presents an opportunity to convey rich visual representations.
\blue{
Both logograms and alphabet languages adhere to these patterns but exhibit distinct preferences.
The complex typeface structure of logograms and their inherent pictorial origins result in more elaborate imagery combinations.
For instance, compared to the French word in Fig.~\ref{fig:corpus} (b5), the Chinese words in Fig.\ref{fig:corpus} (b6) achieve a blending with traditional landscape paintings without spatially rearranging the typefaces.

Additionally, we note distinctions in real-world logo design compared to the formal definition of semantic typography.
In real-world logo design, blending is not uniformly applied to all typefaces. Instead, emphasis is placed on letters in the initial position or those closely related semantically to the imagery ($171/427$ examples).
Moreover, in significant designs, the typeface often has a less direct semantic relationship with the incorporated imagery ($203/427$ examples). This observation aligns with insights from an interview with E2, who remarked, ``\textit{The selection of imagery depends on the brand's story, and the logo's meaning is for users to associate the imagery directly with the brand.}''
}
% select transformed letter flexibly
% semantifc 

% Through corpus analysis, we have uncovered that the typeface structure in semantic typography can be decomposed into different levels of granularity, namely \textit{stroke-}, \textit{letter-} and \textit{word-level}.
% The prominent characteristic of semantic typography lies in the interaction between the typeface and imagery in heterogeneous ways, in which we identified three prevailing patterns: \textit{one-to-one}, \textit{one-to-many}, and \textit{many-to-one mapping}.

%% file: Latex/4_DesignConsideration.tex
\section{Design Consideration}
\label{sec:design-consideration}

% \red{\begin{center}
%     \textit{``Design with low thresholds, high ceilings, and wide walls.''
%     -- Ben Shneiderman \cite{shneiderman2007creativity}}
% \end{center}}

\blue{
The expert interview reveals that personalizing a semantic typographic logo relies on blending specific typefaces (e.g., at different granularity) and imagery (e.g., concrete visual representation), further identified through corpus analysis.
Challenges in user workflow and concern about generative AI highlight using easily accessible images for effective personalization, eliminating the need for intricate text prompts that may not fully capture user intentions.
Guided by Shneiderman's design principle \cite{shneiderman2007creativity} of \textit{``Design with low thresholds, high ceilings, and wide walls,''} our goal is to develop a tool that enables novices to create with accessible materials and interactions (\textbf{D1}), automates complex blend manipulation for professionals (\textbf{D2}), and integrates essential functionalities for a streamlined design process (\textbf{D3}).
The roadmap we derive the design considerations is illustrated in Fig.~\ref{fig:workflow}.
Below, we present the set of design considerations:

\begin{itemize}[leftmargin=6.5mm,label={\textbf{D1.}}]

\item[\textbf{D1.}]
\textbf{Intent-aware design material and interaction.}
We aim to support flexible material selection, allowing the easy switch between typefaces at \uline{different granularities} and the selection of imagery from specific visual representations in a \uline{user-customized image}.

\item[\textbf{D2.}]
\textbf{Facilitate the professional generation process.}
We aim to propose an \uline{automatic blending approach} that supports typefaces at \uline{all levels of granularity}, ensuring harmonious and diverse designs.

\item[\textbf{D3.}]
\textbf{Provide necessary functionalities to support a comprehensive workflow.}
In the pre-generation stage, we will incorporate an ideation module for \uline{brainstormin}g. In the post-generation stage, an evaluation and iteration module will be added to \uline{identify}, \uline{edit}, and \uline{refine} the generated result within the type-imagery spectrum.

\end{itemize}
}

% \begin{itemize}[leftmargin=5.5mm,label={\textbf{D1.}}]

% \item[\textbf{D1.}]
% \textbf{Provide clear and comprehensible brainstorming ideas.}
% Users will access various prospective imageries with clear descriptions to help them find related inspiration, get past design fixation, and unleash their creativity.

% \item[\textbf{D2.}]
% \textbf{Generate harmonious and diverse blends between typeface and imagery.}
% An automatic blend approach will be employed to support type-imagery mapping at various structural granularities, achieving diverse designs.

% \item[\textbf{D3.}]
% \textbf{Improve controllability for a more personalized generation.}
% The generation will be customized with flexible user control to ensure user intents can be incorporated into the result.

% \item[\textbf{D4.}]
% \textbf{Identify the position of the generated result in the type-imagery spectrum.}
% A data-driven discriminator will quantify the type-imagery spectrum, supporting users in balancing between readability and expressiveness.

% \item[\textbf{D5.}]
% \textbf{Enhance refinement and editability for the generated result.}
% Users can adjust their designs in the type-imagery spectrum and edit the element to meet specific requirements after generation.

% \end{itemize}

%% file: Latex/5_TypeDance.tex
\section{TypeDance}
\label{sec:typedance}

Based on the identified design rationales and the highlighted opportunity to address the challenges and concerns in the design workflow, we have developed TypeDance, an authoring system that facilitates personalized generation for semantic typographic logos.
TypeDance comprises five essential components, which closely correspond to the pre-generation, generation, and post-generation stages.
In the pre-generation stage, the \textit{ideation} component communicates with the user to gather comprehensible imageries (\textbf{D1}). The \textit{selection} component allows the user to choose specific design materials, including typeface at various granularities and imagery with particular visual representations (\textbf{D2}).
The \textit{generation} component blends these design materials, utilizing a series of combinable design priors (\textbf{D3}).
After the generation, the \textit{evaluation} component enables the user to assess the current result's position in the type-imagery spectrum (\textbf{D4}).
The \textit{iteration} component empowers user to refine their design by adjusting the type-imagery spectrum and editing each individual element (\textbf{D5}).

\subsection{Ideation}
\label{ssec:ideation}

Thanks to the impressive language understanding and reasoning capabilities of large language models, we can now collaborate with a knowledgeable brain through text. TypeDance takes advantage of Instruct GPT-3 (davinci-002)~\cite{ouyang2022training}  with Chain-of-Thought prompting~\cite{wei2022chain} to generate relevant imagery based on given texts.
To enhance user understanding, the prompting strategy includes the requirement to accompany the imagery with explanations in terms of visual design. This ensures that the answers provided are more interpretable and informative.
For example, when the user provides the keyword ``\textit{Hawaii,}'' TypeDance generates concrete imagery such as ``\textit{Aloha Shirt},'' ``\textit{Hula Dancer},'' and ``\textit{Palm tree}.'' Along with these imagery words, TypeDance also provides explanations like ``\textit{Symbolizes the vibrant culture and traditional dance form of Hawaii}'' for the ``\textit{Hula Dancer}''.
By making this small change, TypeDance provides interpretable explanations and offers users additional background knowledge to enhance their understanding.

\subsection{Selection}
\label{ssec:selection}
\begin{figure}[t]
  \centering
  \includegraphics[width=\linewidth]{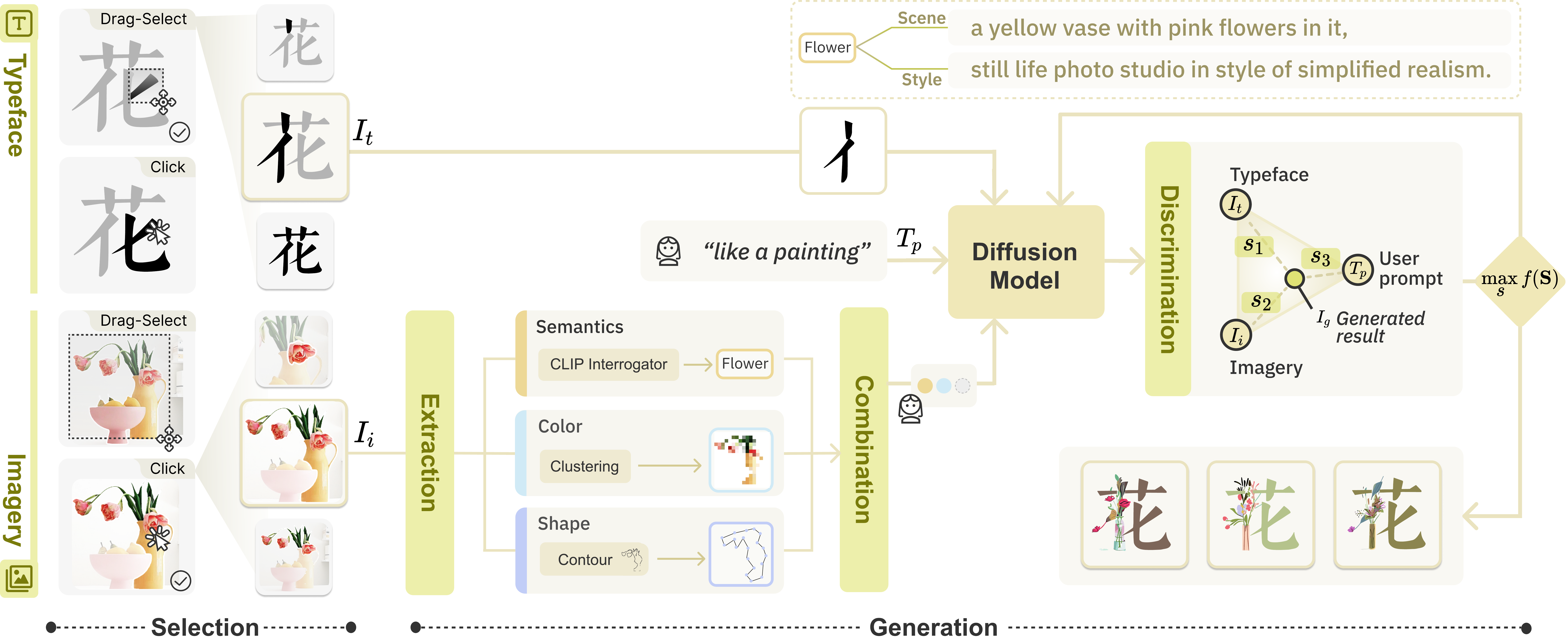}
  \caption{The \textit{Selection} and \textit{Generation} component in the workflow of TypeDance. The \textit{selection} component offers two types of interaction to allow creators to flexibly select typeface $I_{t}$ at different granularity and imagery $I_{t}$ with specific visual representation. These design materials will be injected into the diffusion model in the \textit{generation} component with an optional user prompt $I_{t}$. The discrimination is conducted to ensure the generated result can meet three-dimensional user intent, including $I_{t}$, $I_{i}$, and $T_{p}$.}
  \label{fig:method}
\end{figure}

The selection aims to prepare design materials blended in the following generation.
It encompasses two fundamental components: selecting typeface $I_{t}$ at various granularities and imagery $I_{i}$ with particular visual representations.
We achieve the fine-grained high-fidelity segmentation based on Segment Anything Model~\cite{kirillov2023segany}, which offers user-friendly visual prompts as input, including box and point.
As illustrated in Fig.~\ref{fig:method}, the different characteristics inherent to typefaces and imagery give rise to varying interactions that aim to align with the design rationales.

\noindent\textbf{Typeface Selection.}
Given the text, TypeDance allows creators to select typefaces $I_{t}$ at different granularity, as identified in our formative study, to achieve a more fine-grained and flexible approach to complex design practices.
Instead of being limited to using complete strokes, TypeDance enables creators to select partial regions of a single stroke.
To achieve this, As Fig.~\ref{fig:method} shows, we implemented the drag-select interaction for creators to encompass specific parts of the typeface they need within a designated box.
Compared to the click interaction, the drag-select offers a more explicit way to reveal user intention, which is free from rule-based segmentation that is restricted by a set of predetermined strokes, thus supporting more fine-grained selection.
Moreover, TypeDance offers a combination selection by which creators can select strokes located far apart, as in the case shown in Fig.~\ref{fig:method}.

\noindent\textbf{Imagery Selection.}
Regarding imagery selection, we employ semantic segmentation to extract the visual representation creators require from a cluttered background. However, unlike typeface selection, the nature of imagery selection is more conducive to clicking rather than drag-selection.
As depicted in Fig.~\ref{fig:method}, drag-selection can inadvertently encompass other objects the creator may not need while selecting their desired object. To address this issue, we have implemented a solution where creators can click on individual objects separately, mitigating the problem of unintentional object coverage.
Similarly to typeface selection, TypeDance also supports combination selection, allowing creators to choose multiple objects within the image.

\subsection{Generation}
\label{ssec:generation}

\subsubsection{Input Generation}
This section describes the three inputs required for the generation process. The first input is the selected typeface $I_{t}$, which serves as the origin image for the diffusion model. The second input is the optional user's prompt $T_{p}$, which allows them to explicitly express their intent, such as the specific style they desire. The third input consists of the design factors extracted from the selected image $I_{i}$.

\noindent \textbf{Semantics.}
Textual prompt is an accessible and intuitive medium for creators to instruct AI, which also offers a way to incorporate imagery into the generation process.
However, it is laborious to describe a significant amount of information within the constraints of a limited prompt length.
TypeDance solves this problem by automatically extracting the description of the selected imagery.
Describing the selected imagery involves a text inversion process encompassing multiple concrete semantics dimensions.
One of the prominent semantics is the general visual understanding of a \textit{scene}.
For instance, in Fig.~\ref{fig:method}, the description of the scene is \textit{``a yellow vase with pink flowers.''}
We capture this explicit visual information (object, layout, \textit{etc}.) using BLIP~\cite{li2022blip}, a Vision-Language model that excels in image captioning tasks.
Moreover, the \textit{style} of imagery, especially when it comes to illustrations or paintings, can greatly influence its representation and serve as a common source of inspiration for creators.
The style of the case in Fig.~\ref{fig:method} is \textit{``still life photo studio in style of simplified realism.''}
Such a specific style is derived from retrieving relevant descriptions with high similarity in a huge prompt database.
Therefore, the complete semantics of the imagery include the scene and style.
To enhance interface scalability, we extract keywords from the detailed semantics.
Creators can still access the complete version by hovering over the keywords.

\noindent \textbf{Color.}
TypeDance utilizes kNN clustering~\cite{fix1989discriminatory} to extract five primary colors from the selected imagery. These color specifications are then applied in the subsequent generation process. In order to preserve the semantic colorization relation, the extracted colors are transformed into a 2D palette that includes spatial information. This ensures that the generated output maintains a meaningful and coherent color composition.

\noindent \textbf{Shape.}
The shape of the typeface can take an aesthetic distortion to incorporate rich imagery, as demonstrated in our formative study.
To achieve this, we first leveraged edge detection to recognize the contour of selected imagery.
Then, we sample 20 equidistant points along the contour. 
These points are used to deform the outline of the typeface iteratively, using generalized Barycentric coordinates~\cite{meyer2002generalized}. The deformation occurs in the vector space, resulting in a modified shape that depicts coarse imagery and facilitates guided generation.

These design factors are applied independently during the generation process. Creators have the flexibility to combine these factors according to their specific needs, allowing for the creation of diverse and personalized designs.

\subsubsection{Output Discrimination}
To ensure that the generated result aligns with the creators' intent, TypeDance employs a strategy that filters good results based on three scores. As illustrated in Fig.~\ref{fig:method}, we aim for the generated result $I_{g}$ to achieve a relatively balanced score in the triangles composed of typeface, imagery, and the optional user prompt.
The typeface score $s_{1}$ is determined by comparing the saliency maps of the selected typeface and the generated result. Saliency maps are grayscale images that highlight visually salient objects in an image while neglecting other redundant information. We extract the saliency maps for the typeface and the generated result and then compare their similarity pixel-wise.
The imagery score $s_{2}$ is derived from the cosine similarity between the image embeddings of the input image $I_{i}$ and the generated result $I_{g}$. Similarly, we obtain the prompt score $s_{3}$ by computing the cosine similarity between the image embedding of the generated result $I_{g}$ and the text embedding of the user prompt $T_{p}$. We use the pre-trained CLIP model to obtain the image and text embeddings because of its aligned multi-modal space.
We denote $s_{i} = \{s_{i1}, s_{i2}, s_{i3}\}$, where $i$ represents the $i$-th result at one round of generation.
To filter the results that mostly align with the creators' intent, we use a multi-objective function that maximizes the sum of the scores and minimizes the variance between them. The function is defined as follows:
\begin{equation*}
    \max_{s} f(\mathbf{S}) = \sum_{j=1}^{3} s_{ij} - \lambda \cdot \sigma(\mathbf{S}),
\end{equation*}
where $\mathbf{S}$ is the score set of all generated results, and $\sigma(\mathbf{S})$ calculates the variance of the scores. The $\lambda$ is a weighting factor used to balance the total score and variance, which is empirically set as $0.5$.
Based on this criteria, TypeDance displays the top $1$ result on the interface each round and regenerates to obtain a total of four results.

\subsection{Evaluation}
\label{ssec:evaluation}

Once the design is generated, evaluating a design's compliance with recognized visual design principles is crucial for its completion~\cite{petrie2004tension, chilton2019visiblends}.
Semantic typographic logos require a balance between the typeface's readability and the imagery's expressiveness, presenting a challenging tradeoff.
Designers often rely on the feedback from participants to validate their designs. Typdance offers a data-driven instant assessment before gathering participants' reactions.

To assist users in determining the position of their current work on the type-semantic spectrum, TypeDance utilizes a pre-trained CLIP model~\cite{radford2021learning} that provides objective judgment supported by data.
By distilling knowledge from a vast dataset of 400 million image-text pairs of CLIP model, TypeDance can quantify the similarity between typeface and imagery on a scale ranging from $[0, 1]$, with the sum equal to $1$.
To enhance intuitive understanding, TypeDance translates the neutral points from $0.5$ to $0$ and normalizes the distance between the similarity and the neutral point from $0.5$ to $1$.
Users can determine the degree of divergence from the neutral point for their currently generated result. A value of $0$ signifies neutrality, indicating that the generated result favors both the typeface and the imagery. Conversely, higher values indicate a greater degree of divergence towards either the typeface or the imagery, depending on the specific direction.

\subsection{Iteration}
\label{ssec:iteration}
Although iteration is throughout the process, we identified three main iteration patterns.

\subsubsection{Regeneration for New Design.}

Previous tools often allow creators to delete unsatisfactory results, but few aim to improve the regenerated performance to meet creators' intent. 
To address this challenge, TypeDance utilizes both implicit and explicit human feedback to infer user preferences. Inspired by FABRIC~\cite{von2023fabric}, we leverage users' reactions toward generated results as a clue to user preference. 
By analyzing positive feedback from preserved results and negative feedback from deletions, the generative model in TypeDance dynamically adjusts the weights in the self-attention layer. This iterative incorporation of human feedback allows for the refinement of the generative model over time. 
Additionally, TypeDance provides users with more explicit ways to make adjustments, including a textual prompt and a slider that enables users to control the balance between typeface and imagery.

\subsubsection{Refinement in Type-imagery Spectrum.}

TypeDance provides a refined approach to iteratively refine their designs along the type-imagery spectrum.
In addition to the quantitative metric identified in the evaluation component, TypeDance allows creators to make precise adjustments using the same slider.
As shown in Fig.~\ref{fig:interface} (c), we divide the distance between the neutral point and the typeface and imagery into 20 equal intervals, each representing $0.05$. These small interpolations preserve the overall structure and allow for incremental adjustments.
By dragging the slider, creators can set their desired value point between typeface and imagery, achieving a balanced aesthetic.

Specifically, to prioritize imagery, we begin with the current design as the initial image and inject it into the diffusion model with the strength set to the desired value point. This process pushes the generated image toward the imagery end, resulting in a more semantically rich design.
Conversely, to emphasize the typeface, we utilize the saliency map of the typeface in the pixel space to filter out the relevant regions of the image. This modified image is fed into the generative model to ensure a smooth transition towards the typeface end.

\subsubsection{Editability for Elements in the Final Design.}
A more fine-grained adjustment is required to make nuanced changes to an almost satisfactory result, such as deleting an element in the generated result. In order to achieve this level of editability, we convert the image from pixel to vector space. Hence, creators gain the ability to manipulate each individual element in their design, allowing them to remove, scale, rotate, and change colors as needed.

\section{Interface Walkthrough}
\label{sec:walkthrough}

In TypeDance, the components that follow the design workflow are organized in a cohesive U-shaped layout on the interface, with the main canvas at the center, as depicted in Fig.~\ref{fig:interface}.
This design aims to facilitate seamless navigation for creators, eliminating the need for constant switching between different components.
Following the user workflow, we demonstrate how TypeDance creates visually appealing designs.
Alice is a graphic designer who wants to create a series of postcards for the four seasons using imaginative semantic typographic illustrations.
Starting her creative journey, Alice begins with the character ``\raisebox{-.15\height}{\includegraphics[height=0.26cm]{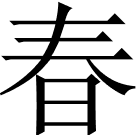}},'' which means \textit{``Spring''} in Chinese.

\subsection{Pre-generation stage}

\subsubsection{Seeking design ideas.}
To gather design inspiration, Alice first types the \textit{``Give me some ideas about spring''} in the {\ttfamily{IDEATION}} \raisebox{-.15\height}{\includegraphics[width=0.34cm]{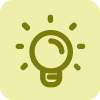}} and click the {\ttfamily{[Brainstom]}} button.
TypeDance generates a list of ideas like \textit{``Bird in Nest''} and \textit{``Blooming Flowers''}.
She hovers over these ideas, and their corresponding explanations pop up. 

\subsubsection{Preparing design materials.}
Alice types the character ``\raisebox{-.15\height}{\includegraphics[height=0.26cm]{figures/icons/characters-spring.png}}'' in the {\ttfamily{TYPEFACE}} \raisebox{-.15\height}{\includegraphics[width=0.38cm]{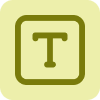}} and selects the font type as \textit{``Mincho''}. 
She uses the drag-selection to choose the lower part of the character, represented as ``\raisebox{-.15\height}{\includegraphics[height=0.24cm]{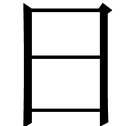}},'' and clicks {\ttfamily{[\checkmark]}} button to confirm her selection.
Consequently, the rest of the typeface appears on the central canvas.
Next, she opens another tab to search for images using the keyword \textit{``Blooming Flowers in Spring''}.
After selecting an appropriate image, she uploads it to the {\ttfamily{IMAGERY}} \raisebox{-.15\height}{\includegraphics[width=0.38cm]{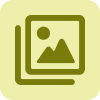}} by clicking to select the window object within the image. Upon doing so, the selected window object is highlighted. To finalize her selection, Alice clicks the {\ttfamily{[\checkmark]}} button.

\begin{figure}[t]
  \centering
  \includegraphics[width=\linewidth]{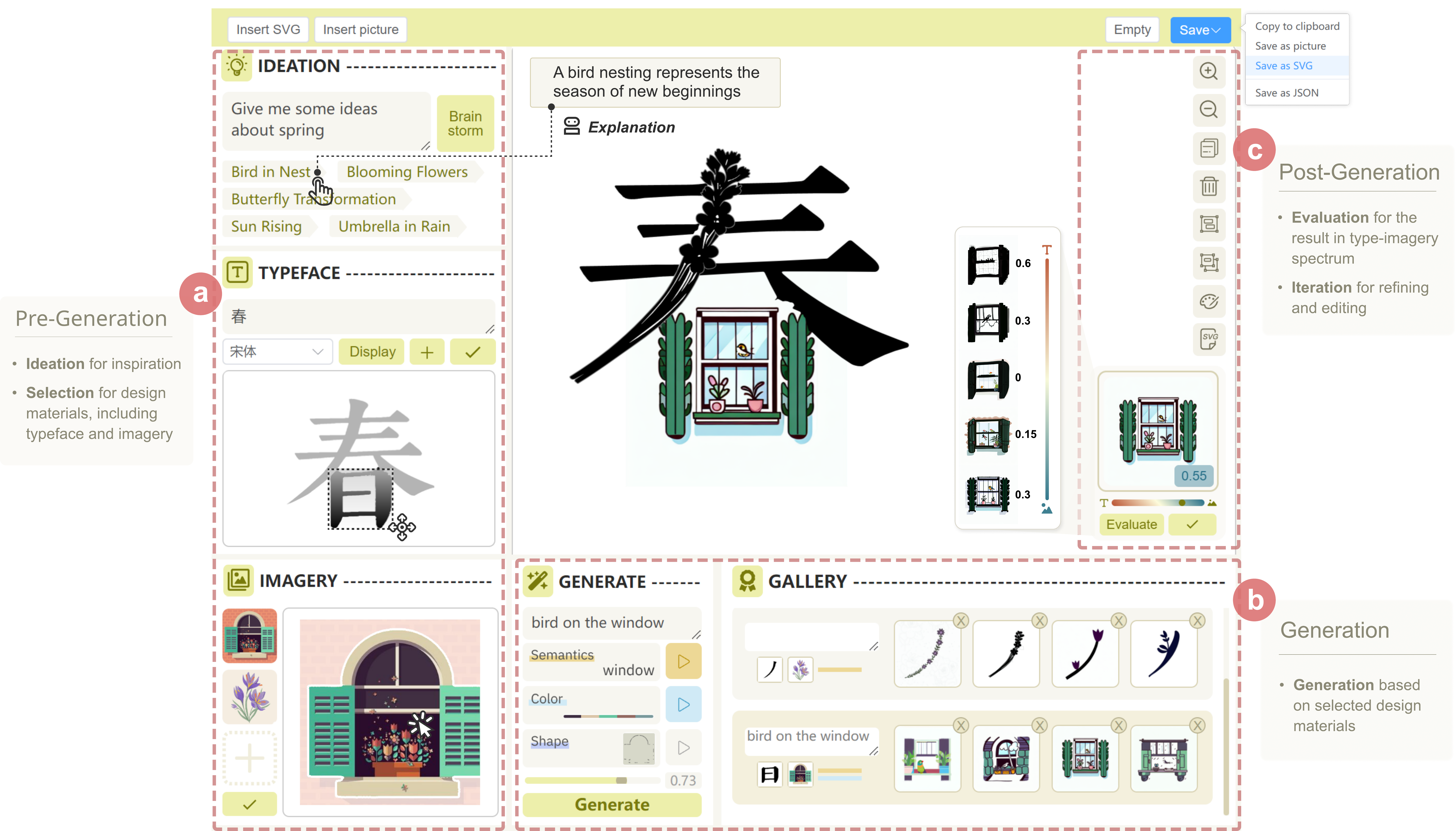}
  \caption{The interface of TypeDance, with a creator engaging in semantic typographic design. (a) In pre-generation, creator brainstorms for ideas and selects typeface and imagery as design materials. (b) During generation, creator sets generation options along with a prompt to personalize the design. (c) In post-generation, the creator evaluates and refines the design in the type-imagery spectrum.
  }
  \label{fig:interface}
\end{figure}

\subsection{Generation stage}

\subsubsection{Blending typeface and imagery.}
Alice browses the extracted design factors in the {\ttfamily{GENERATE}} \raisebox{-.15\height}{\includegraphics[width=0.38cm]{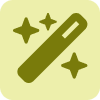}} and selects the {\ttfamily{[semantic]}} and {\ttfamily{[color]}} options.
She leaves the strength at its default setting of $0.75$ for her first attempt.
To add more imagery of spring, she looks back to the {\ttfamily{IDEATION}} and notices the \textit{``Bird in Nest''}.
she proceeds to input the prompt \textit{``bird on the window''} in the {\ttfamily{GENERATE}} and submit the generation. 
After a brief 15-second wait, the results are presented in the {\ttfamily{GALLERY}} \raisebox{-.15\height}{\includegraphics[width=0.38cm]{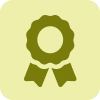}}.

\subsubsection{Regenerating with appropriate strength.}
Recognizing the generated result is closer to the typeface, she deletes unwanted results and adjusts the design factor strength to $0.86$ using the slider.
In the subsequent round, she finds a desirable result and clicks it.
The chosen design is then displayed in the central canvas. 

\subsection{Post-generation stage}

\subsubsection{Evaluating and Refining the generated result}
To assess its legibility, Alice navigates to the right side of the canvas and clicks the {\ttfamily{[Evaluation]}} button in the {\ttfamily{EVALUATION}}.
The current position of the result is situated on the imagery side of the slider with a value of 0.55. However, aiming to explore positions more aligned with the typeface side, she drags the slider to the left.
After several trials, Alice obtains a series of results, as shown in Fig.~\ref{fig:interface}.

\subsubsection{Editing and Exporting.}
Alice repeats the process to blend a stroke ``\raisebox{-.15\height}{\includegraphics[height=0.26cm]{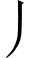}}'' with the floral imagery.
She converts them into SVG format and modifies the color of the flowers.
Finally, Alice exported the design by clicking the {\ttfamily{[save]}} button on the top-right corner of the interface.

%% file: Latex/6_Evaluation.tex
\section{Evaluation}
\label{sec:evaluation}

To test the effectiveness of TypeDance, we conducted \blue{a baseline comparison and} user study.
Our primary objective was to evaluate the performance of generated results and the usability of TypeDance, and explore how each component could potentially address pain points in their design workflows.
Additionally, we delved into the limitations of the tool and identified opportunities for improvement.

\subsection{Baseline Comparison}

\begin{figure}[t]
  \centering
  \includegraphics[width=0.98\linewidth]{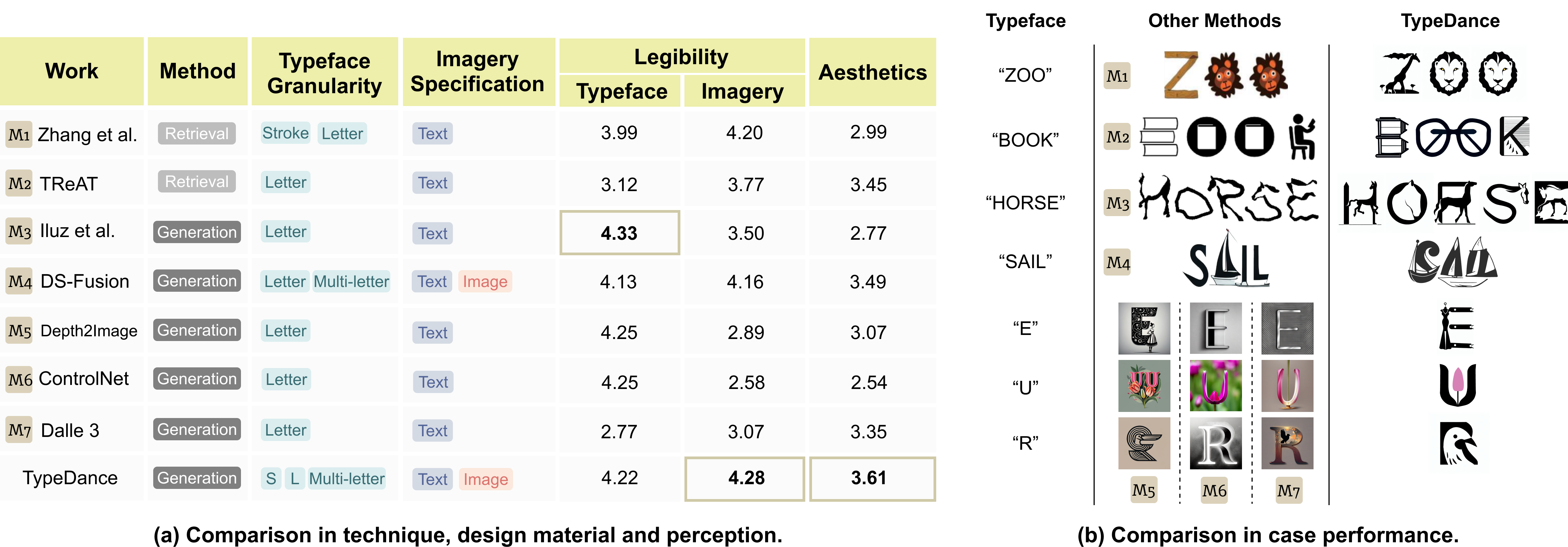}
  \caption{
  \blue{TypeDance \textit{vs.} baselines: (a) comparison in technique, design material, and perception, and (b) comparison in case performance.}
  }
  \label{fig:baseline}
\end{figure}

% method list
\blue{
We conducted a comparison with seven alternative methods: Zhang et al. (M1)\cite{zhang2017synthesizing}, TReAT (M2)\cite{tendulkar2019trick}, Word as Image (M3)\cite{IluzVinker2023}, DS-Fusion (M4)\cite{tanveer2023ds}, Depth2Image (M5)\cite{mou2023t2i}, ControlNet (M6)\cite{zhang2023adding}, Dalle 3 (M7)\cite{betker2023improving}.
For a comprehensive evaluation, we assessed these methods from both technical and perceptual perspectives, as depicted in Fig~\ref{fig:baseline}. Given that most works are not open-source, an overall comparison using the same case was not feasible. Instead, we randomly sampled three cases from each method and utilized TypeDance to recreate them with the same text content and imagery, taking a one-to-one comparison with each method.
The full cases are listed in the supplemental material.
The perception study involved an online questionnaire with the participation of 50 individuals.
We shuffled the appearance sequence of all logos and provided no hints about the original typeface or imagery used, and the scores were recorded on a 5-point Likert scale.
}
% 图上提示1~5

\subsubsection{Technique Difference.}
\blue{
Artistic typography and TReAT operate on a retrieval-based approach, which limits their ability to generalize to cases significantly different from the collected templates. In contrast, other methods adopt a generation-based approach, offering some alleviation to this limitation. However, most of them are restricted to letter-level blending, whereas TypeDance excels by supporting blending at all typeface granularity.
Regarding the interaction to specify imagery, the majority of methods rely on text, either by retrieving relevant templates from a corpus or guiding the generative model. Notably, DS-Fusion and TypeDance stand out as they support using images to assign specific visual representations. It's worth noting, though, that DS-Fusion necessitates users to provide a small image dataset of around 20 images for fine-tuning the model, a process taking approximately 1.5 hours using a desktop with Nvidia GeForce RTX 3090.
}

\subsubsection{Perception Study.}
\blue{
We used two primary metrics to assess the performance of these methods: one focused on the legibility of typeface and imagery, and the other on aesthetics. As illustrated in Fig.~\ref{fig:baseline}, TypeDance outperforms other methods in both aesthetics score and legibility of imagery.
Word as Image achieves the highest score in the legibility of typeface, but its imagery is comparatively challenging to recognize. Similarly, many methods exhibit this imbalance, excelling in one representation while compromising the other.
In contrast, TypeDance maintains a stable performance with commendable aesthetic recognition.
}

\subsection{User Study}
\subsubsection{Participants}
Both designers and general users (\blue{11} females and \blue{7} males, aged \blue{19}-34) are invited to obtain different feedback.
\blue{Nine} participants (P1-\blue{P9}) are novice users interested in semantic typography art without formal design training.
The remaining \blue{nine} are graphic designers (E1-\blue{E9}) with professional design education with more than three years of experience in semantic typographic logos.
All participants have tried AI tools like Midjourney before.
They accessed TypeDance through web browsers, utilizing a combination of online and offline modes. As a token of appreciation, each participant received a \$30 gift card upon completing the study.

\begin{figure}[t]
  \centering
  \includegraphics[width=\linewidth]{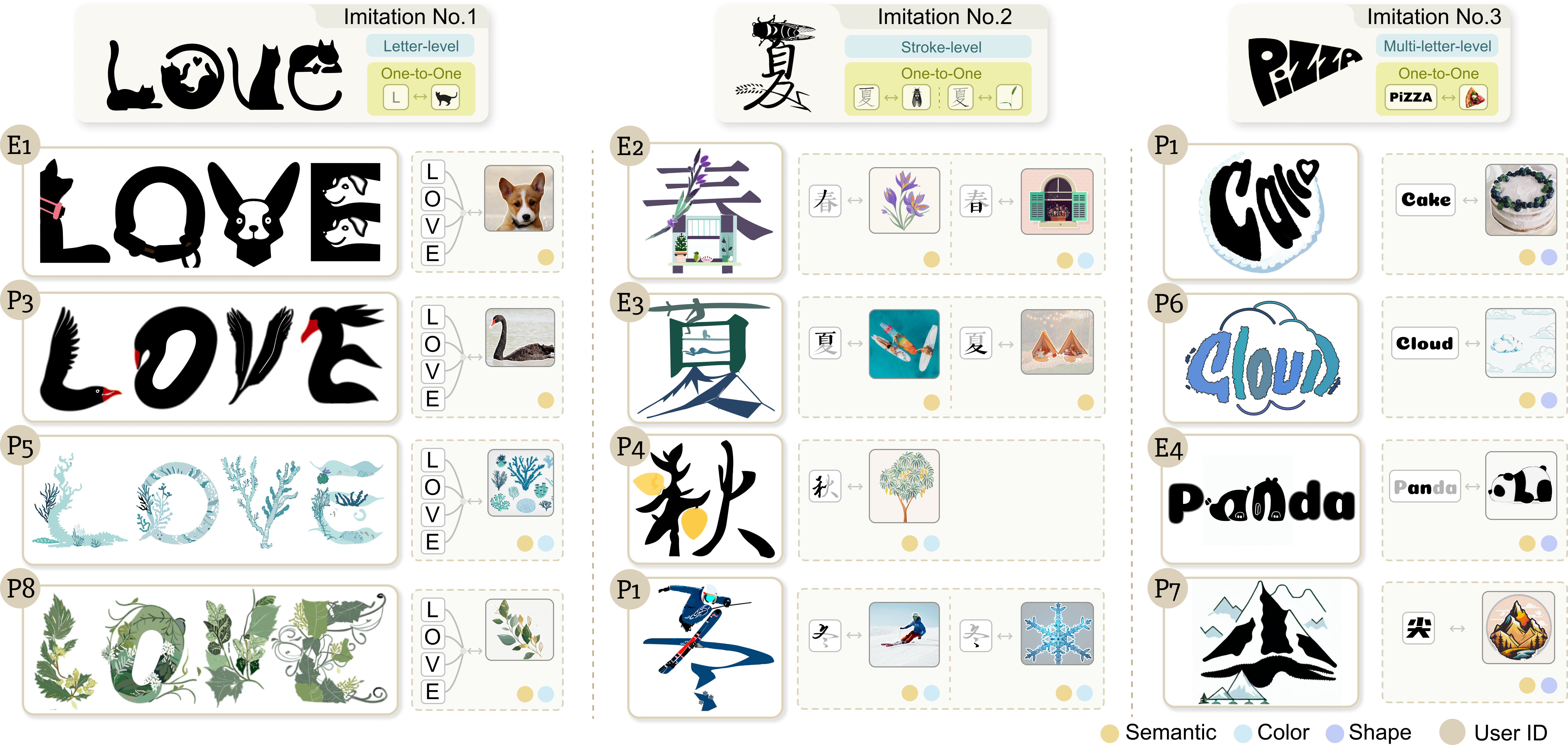}
  \caption{The first task of evaluation is imitation. In this task, users are required to choose two out of three references and imitate their style. Part of the design outcomes created by participants via TypeDance are shown, with annotations indicating the necessary design materials, including typeface and imagery, and the design patterns in their designs.}
  \label{fig:imitation}
\end{figure}

\subsubsection{Tasks and Procedure}
\begin{figure}[t]
  \centering
  \includegraphics[width=\linewidth]{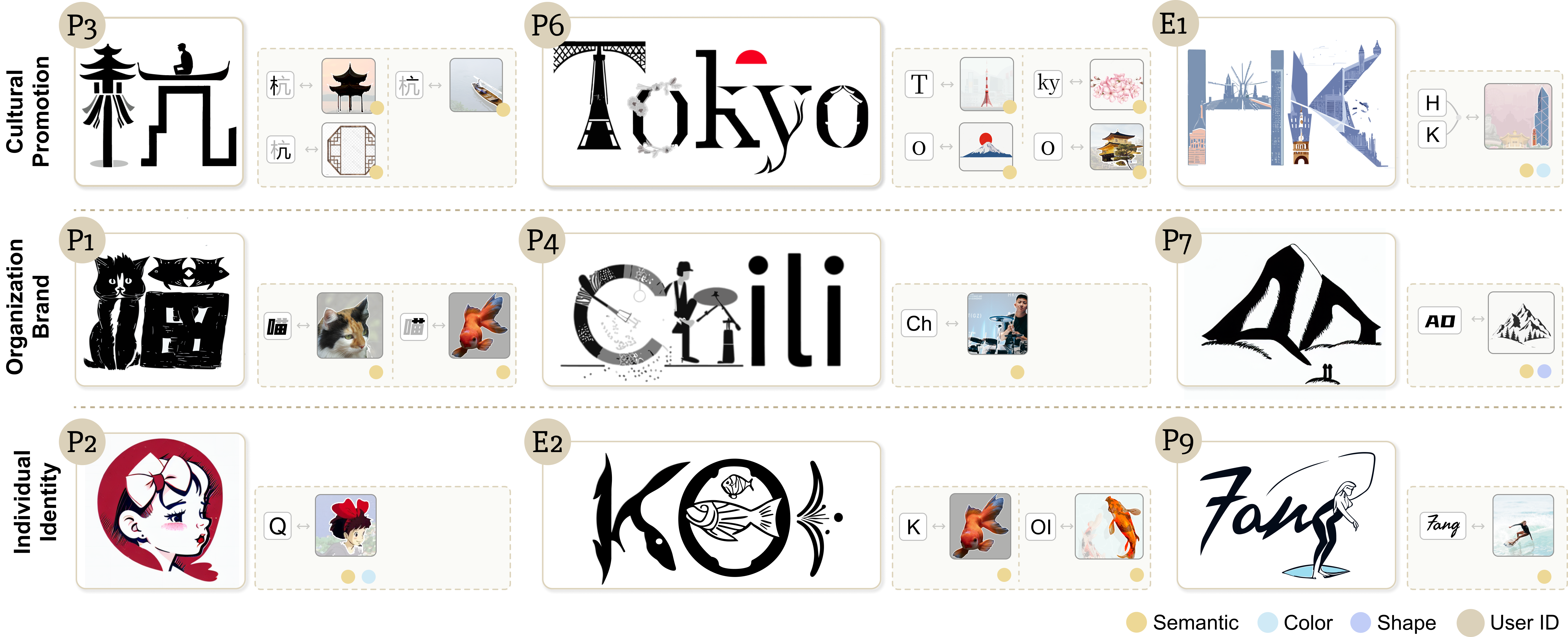}
  \caption{
  % The second task of evaluation is creation, which requires users to pick a scenario from cultural promotion for their favorite city, organizational branding, or personal identity to create their own logo. Part of the design outcomes created by participants via TypeDance are shown, with annotations indicating the necessary design materials, including typeface and imagery, and the design patterns in their designs.
  \blue{The second evaluation task is creation. For participants without a concrete creation goal, we provided open-ended topics as inspiration.} Part of the \blue{open-ended topics} design outcomes created by participants via TypeDance are shown, with annotations indicating the necessary design materials, including typeface and imagery, and the design patterns in their designs.
  }
  \label{fig:creation}
\end{figure}

Drawing inspiration from the design process outlined by Okada et al.~\cite{okada2017imitation}, \blue{which contains two essential stages, namely imitation and creation,} we crafted two tasks accordingly to align the natural and progressive design challenges. The comprehensive procedure is outlined as follows:

\begin{itemize}[leftmargin=6.5mm]
\item Briefing (20 minutes).
We presented several examples of semantic typographic logos from our collected corpus to introduce the topic. To facilitate a swift transition for participants into their roles as creators, we inquired about their preferred design and asked them to envision the steps required to accomplish such a design. Next, we introduced TypeDance using the example depicted in Fig.~\ref{fig:method}. Participants were then encouraged to independently explore TypeDance for $5$ minutes to gain further familiarity with its functionalities.

\item Task1: Imitation (25 minutes).
% We chose three design references that encompass the design patterns identified in the formative study.
Participants are instructed to select two of three design references and replicate their styles in their own designs.
Reference $1$ encourages participants to incorporate imagery representing their passions within the typeface \textit{``love''} at the letter-level.
Reference $2$ endeavors to integrate as much imagery as possible, depicting the four seasons within the typeface ``\raisebox{-.15\height}{\includegraphics[height=0.26cm]{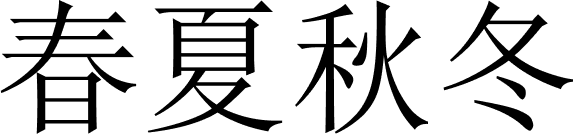}}'' at the stroke-level.
Reference $3$ focuses on the shape distortion of typefaces at the multi-letter level.
Some user design outcomes are shown in Fig.~\ref{fig:imitation}.

\item Task2: Creation (20 minutes).
\blue{Participants are encouraged to explore and create their own designs freely.
For those without a specific creative direction, we provide open-ended topics drawn from our collected corpus (cultural promotion, organizational branding, and personal identity) to inspire ideas.} Thinking aloud during the creation process is encouraged, and some user design outcomes are shown in Fig.~\ref{fig:creation} and Fig.~\ref{fig:creation2}.
% We encouraged participants to create their own designs freely.
% To alleviate the problem of lacking of motivation, we provided three open-ended topics (cultural promotion for their favorite city, organizational branding, and personal identity) to them when they needed it.

\item Interview (20 minutes).
In the end, each participant completed a questionnaire with a 5-point Likert Scale.
The questionnaire focused on three aspects of TypeDance: 1) the satisfaction of the generated outcome, 2) the usability of TypeDance system, and 3) the functionality of each individual component within the system.
In addition, we conducted a semi-structured interview with each participant to collect their feedback on the design process.
\end{itemize}

\subsection{Results Analysis}

\subsubsection{Satisfaction of Generated Outcome.}
All Participants found that the generated outcome effectively blends both the information of the selected typeface and imagery \blue{($MEAN=4.78, SD=0.43$)}, and the majority of them \blue{($MEAN=4.17, SD=0.62$)} agree the outcome can achieve a visually harmonious effect.
Additionally, Over half of the participants \blue{($MEAN=4.06, SD=0.73$)} acknowledged that the generated outcomes were diverse.
Their feedback supports that TypeDance is capable of achieving a natural blend and providing diverse results, which aligns with the second design consideration (\textbf{D2}) defined in Sect.~\ref{sec:design-consideration}.

\begin{figure}[t]
  \centering
  \includegraphics[width=\linewidth]{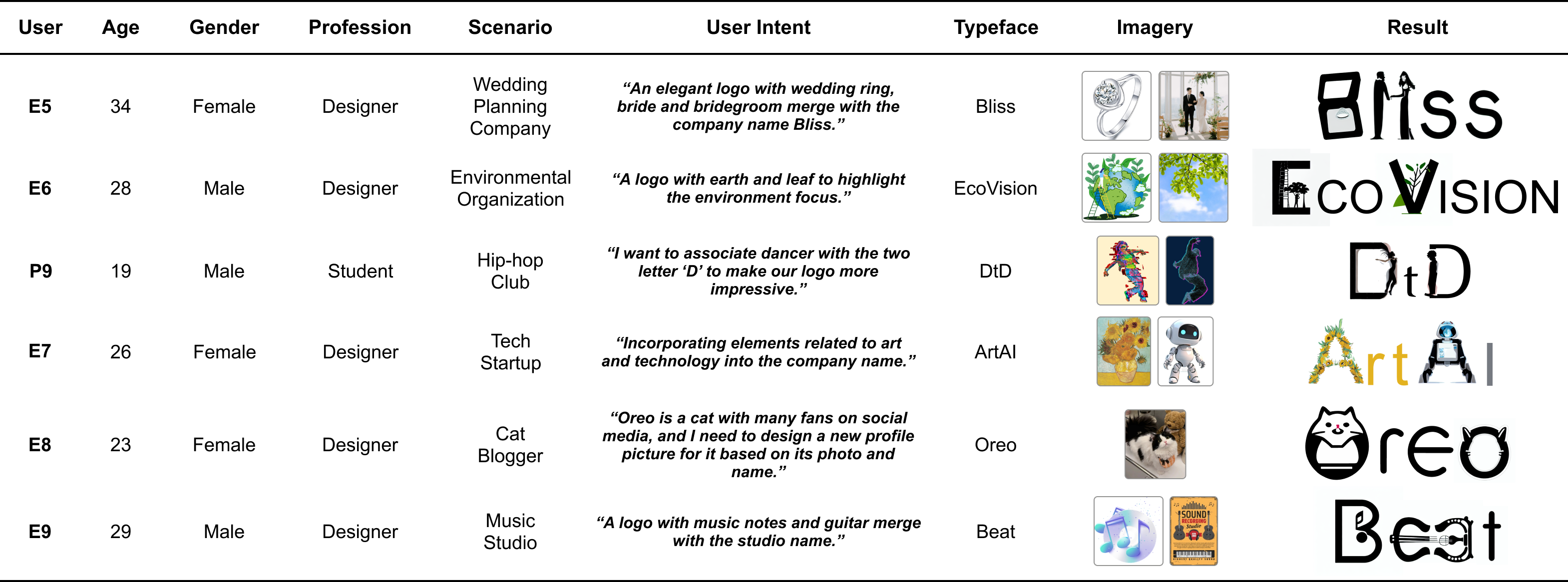}
  \caption{
  % The second task of evaluation is creation, which requires users to pick a scenario from cultural promotion for their favorite city, organizational branding, or personal identity to create their own logo. Part of the design outcomes created by participants via TypeDance are shown, with annotations indicating the necessary design materials, including typeface and imagery, and the design patterns in their designs.
  \blue{In the creation task, some of the participants used TypeDance to fulfill their specific creative needs. For those participants, their personal information, scenario, and creation intent are recorded. Additionally, the necessary design materials, including typeface and imagery, along with the design outcomes are shown in the table. }  
  }
  \label{fig:creation2}
\end{figure}

\begin{itemize}[leftmargin=5.5mm]
    \item \textbf{Preservation.} 
    % It was observed that participants recognized the ability of TypeDance to blend different typefaces and imagery, even when the shape of the typeface significantly differed from the imagery. 
    % The majority of participants (N=13) expressed that the generated results were \textit{``beyond their expectation''} and \textit{``innovative.''} They found that TypeDance was capable of producing reasonable results that effectively combined both typeface and imagery. One participant mentioned, \textit{``I initially didn't see any relation between the swan and the letter ``E,'' but the result showed that they could be combined in a way that is visually pleasing (P3, Fig.~\ref{fig:imitation}).''} This indicates that TypeDance has the ability to comprehend the visual representation of imagery and flexibly adapt to typefaces that may differ with the imagery significantly, as mentioned by another participant who said, \textit{``the coral varies in each letter, and it seems that TypeDance automatically selects the most appropriate type of coral to fit the targeted letter (P5, Fig.~\ref{fig:imitation}).''}
    The majority of participants (\blue{11/18}) expressed that the generated results were \textit{``beyond their expectation''} and \textit{``innovative.''} They found that TypeDance was capable of producing reasonable results that effectively combined both typeface and imagery. As mentioned by P3, \textit{``I initially didn't see any relation between the swan and the letter `E,' but the result showed that they could be combined in a way that is visually pleasing (P3, Fig.~\ref{fig:imitation}).''}
    
    \item \textbf{Harmony.}
    % Participants (16/18) agreed that the generated results exhibited aesthetical harmony. TypeDance successfully maintained the legibility of the typeface while enhancing the visual appeal by incorporating imagery that \textit{``aligned with the skeleton of the text (P1, P4, Fig.~\ref{fig:imitation}).''}
    % Some results demonstrated design expertise that went beyond simply aligning the stroke of the typeface with the imagery.
    % As E1 commented, \textit{``I never thought that AI could understand and utilize design principles like negative space in the design, but it clearly did (Fig.~\ref{fig:imitation}).''}
    % In that case, Typedace integrated the dog into the typeface by filling the empty space in the letter \textit{``E''}.
    % Moreover, P1 and P8 highlighted the importance of visual detail, such as the vines surrounding the leaves and the cream around the cake, in enhancing the \textit{``overall aesthetic harmony (Fig.~\ref{fig:imitation}).''}
    Participants (\blue{16/18}) agreed that the generated results exhibited aesthetical harmony. TypeDance successfully maintained the legibility of the typeface while enhancing the visual appeal by incorporating imagery that \textit{``aligned with the skeleton of the text (P1, P4, Fig.~\ref{fig:imitation}).''}

    \item \textbf{Diversity.}
    Over half of the participants agreed that the generated results were diverse (\blue{14/18}). Some participants \blue{(N=4)} emphasized the importance of obtaining alternative designs in practice, commented that\blue{\textit{``Though I have achieved a satisfactory result, I still want to regenerate to see more interesting results (P2, Fig.~\ref{fig:creation}; E7, Fig.~\ref{fig:creation2}).''}}

\end{itemize}

\blue{
\noindent
In terms of \textit{preservation}, general users exhibited a lower sensitivity than designers in recognizing the typeface and imagery. In contrast, designers could swiftly perceive the content and showed a tendency to export potential designs to advanced tools for further preservation enhancement.
Despite differing levels of design expertise, both novice users and designers demonstrated similar scores in terms of the \textit{harmony} and \textit{diversity} of the generated results.
Besides perceptual harmony, designers identified more artistic effects.
As E1 commented, \textit{``I never thought that AI could understand and produce negative space (Fig.~\ref{fig:imitation}).''}
In that case, TypeDance integrated the dog into the typeface by filling the empty space in the letter \textit{E''}.
During the creation process, all participants experimented with different combinations of design priors to achieve more diversified results. Interestingly, color was more frequently employed than shape, while semantics were consistently selected without specifying a text prompt.
}

\subsubsection{Usability of System}
The user study indicated that most participants \blue{($MEAN=4.39, SD=0.67$)} found TypeDance to maintain workflow integrity in the design process. Additionally, a majority \blue{($MEAN=4.33, SD=0.77$)} expressed satisfaction with the flexibility of blending different granularities of typeface and imagery. In terms of controllability during the generation process and editability of the generated result, more than half of the participants \blue{(N=12)} agreed that TypeDance provided satisfactory control and editability options.
These features align with the design considerations of customization and post-editability (\textbf{D3} and \textbf{D4}) defined in Sect.~\ref{sec:design-consideration}.

\begin{itemize}[leftmargin=5.5mm]
    \item \textbf{Integrity.} 
    Most participants \blue{(N=$10$)} strongly agreed the complete workflow has been instantiated within TypeDance.
    A participant highlighted, \textit{``I don't need to switch between different platforms to finish a design (E2, Fig.~\ref{fig:creation}).''}
    % Some General users mentioned \textit{``feel like hiring a professional designer (P4, Fig.~\ref{fig:creation}; P7, Fig.~\ref{fig:imitation}).''}
    % For participants who are general users who lack design experience, TypeDance was praised for its step-by-step guidance in the design creation process. 
    % Participants mentioned that they only needed to follow the components in the interface and provide requirements at each step, making it \textit{``feel like hiring a professional designer (P4, Fig.~\ref{fig:creation}; P7, Fig.~\ref{fig:imitation}).''}

    \item \textbf{Flexibility.}
    Half of the participants \blue{(N=9)} strongly agreed with the flexibility provided by TypeDance to personalize their designs. 
    Most participants \blue{(N=15)} experimented with more than two types of typeface granularity in their designs.
    E2 pointed out, \textit{``I can easily select a single stroke that overlaps with other strokes in the typeface.''}
    % Besides the stoke and letter selection, some participants found that TypeDance performed well when selecting multiple letters for their designs (P4, P9, Fig.~\ref{fig:creation}). 

    \item \textbf{Controllability.}
    More than half of the participants \blue{(N=12)} agreed that TypeDance provides a high level of controllability.
    They found that the generated results were able to accurately \textit{``reflect the selected imagery'' and ``adhere to the chosen shape''. }
    % Participants mentioned that design factors could be easily transferred from the selected imagery. One participant stated, \textit{``I can decide whether the logo should follow the color or shape, or if both of them should be incorporated into my design (E1, Fig.~\ref{fig:creation}).''} 

    \item \textbf{Editability.}
    The post-editability of Typdance was strongly agreed upon by half of the participants \blue{(N=8)}. Several participants (N=3) expressed their desire for a generative tool that not only generates designs once but also provides the ability to make adjustments and rectify the results.
   %  Refinement of results in the typeface-imagery spectrum was deemed practically useful for semantic typographic design by all experts, e.g., \textit{``The simplification of imagery to make it more like the typeface is a common approach in design, now it can be adjusted in a progressive interpolation.''}
   % The participants unanimously agreed on the SVG conversation and editability. For instance, P4 and P6 changed the color of certain elements in their designs to make them more vibrant(Fig.~\ref{fig:creation}).  
    
\end{itemize}

\blue{
\noindent
All participants widely recognized the workflow \textit{integrity} of TypeDance, with different perspectives from designers and general users.
Designers valued it for integrating essential functionalities that typically require switching between various platforms in the traditional workflow, while general users praised TypeDance for allowing them to sequentially follow the components in the interface to finish a design.
Logo demands high customization with their special property of revealing identity.
The option to select imagery from personal photos adds a personalized touch, surpassing the resources available in a shared community. 
Both designers and novices emphasized the ability to \textit{control} and draw inspiration from the real world with specified visual representation, color, and shape. 
This feature is especially crucial in some scenarios, \textit{e.g.}, ``\textit{designing a city logo.}''

\noindent
The gap between flexibility and editability demonstrates the different expectations from designers and general users.
General users demonstrated less interest in experimenting with different granularities of typeface, predominantly utilizing letter-level blending. 
Designers, on the other hand, highly praised this function as it allows them to segment various parts of the typeface or even combine across different granularities.
After gaining the generated results, general users express satisfaction with changing colors or deleting elements (P4 \& P6, Fig.~\ref{fig:creation}).
Designers find delight in the refinement function, as E5 notes ``\textit{it simulates the real design process where imagery is progressively simplified or details are added to the typeface (E5, Fig.~\ref{fig:creation2}).}''
They also expressed a desire for more advanced editing functions, such as bezier curves, to fine-tune shapes."
}

\begin{figure}[t]
  \centering
  \includegraphics[width=\linewidth]{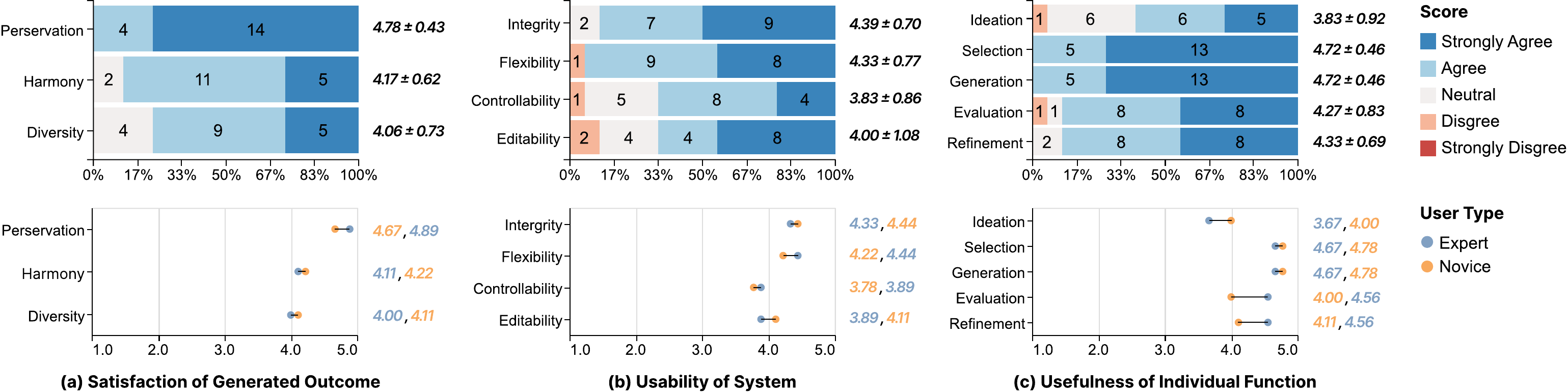}
  \caption{Users ratings for TypeDance, including (a) satisfaction of generated outcome, (b) usability of the system, and (c) usefulness of individual function.
  \blue{
  The above three stacked bar charts show the overall user rating for different indices of TypeDance, while the bellowing three range dot plots illustrate the distinct preferences between experts and novices, showcasing the mean rating values for these two user groups.
  }}
  \label{fig:rating}
\end{figure}

\subsubsection{Usefulness of Individual Functions.}
Participants also provided evaluations for each component within the TypeDance system. The selection and generation components received unanimous agreement from all participants, with high and comparable scores.
The coherence between selecting the typeface and imagery and considering design factors, such as preparing design materials, had a direct impact on the scores of the selection and generation components. 
E3 stated, \textit{``It saves much time for me to select the desired typeface and adjust the bezier curves to create shapes resembling specific objects, like a dog.''}
They also appreciated the diverse range of results offered by the system, which they found crucial for the design process (N=3).
For the pre-generation, In terms of ideation, more than half of the participants \blue{(N=11)} agreed that the concepts provided during the pre-generation phase are \textit{`` helpful to extend imagination''} and \textit{``the explanation makes sense for me''.}
During the post-generation stage, the scores for evaluation and refinement were comparable due to the cohesive nature of the operations. Some participants (N=4, including E1, E3, and E4) expressed a particular satisfaction with these two components, as TypeDance achieves a \textit{``recognization the similarity between typeface and imagery''} and \textit{``a more fine-grained adjustment that is independent of the generation''.}
These post-generation tools are \textit{``especially suitable for semantic typography design,''} said by E1.

\subsection{Limitation}
\label{ssec:limitation}
In response to the issues encountered by users while using TypeDance, we identified the main limitations of the current system from \blue{three} dimensions. 
% These include the balance between the diversity and consistency of generated results and the impact of typeface complexity on the legibility of the generated results. 

\begin{figure}[t]
  \centering
  \includegraphics[width=0.9\linewidth]{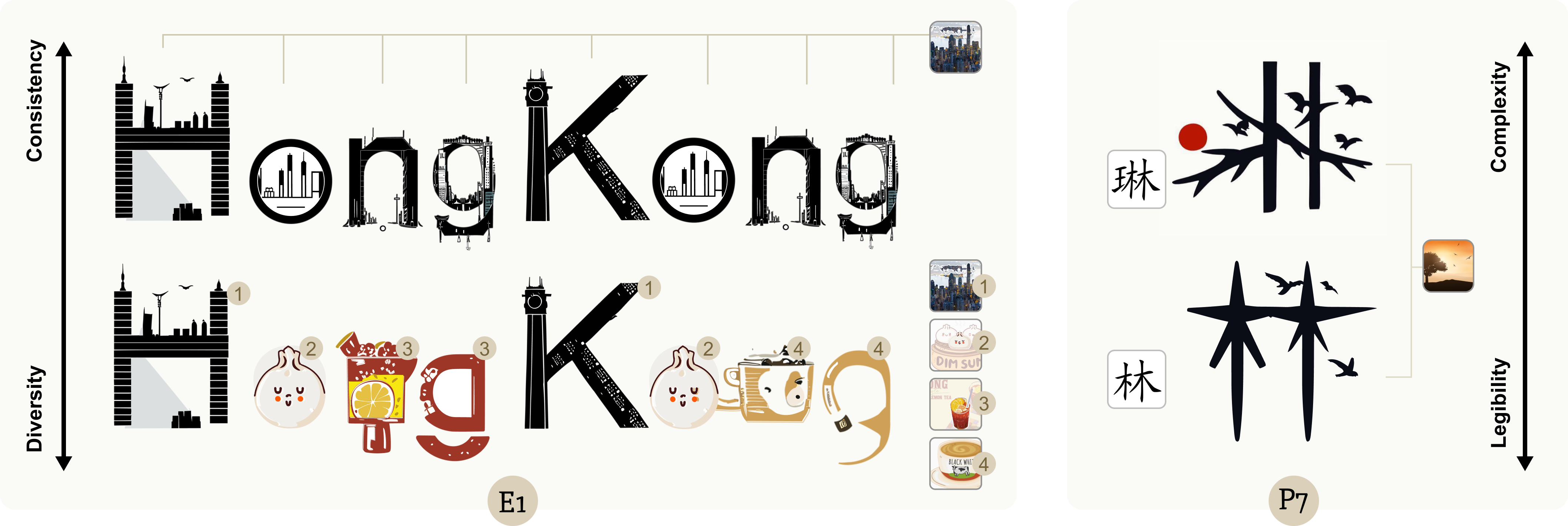}
  \caption{The two tradeoffs revealed in participants' cases. The figure on the left shows the tradeoff between imagery Diversity and style consistency, where the increasing number of imagery will lead to style variations. The figure on the right side shows the tradeoff between the typeface complexity and legibility, where the increased typeface complexity may lead to a potential decrease in the legibility of the results.}
  \label{fig:limitation}
\end{figure}

\subsubsection{Trial and Error in Selecting Typeface and Imagery.}
\blue{
Although the current TypeDance allows creators to select and blend flexibly, facilitating quick generation, participant feedback suggests that trial-and-error with heterogeneous mapping between typeface and imagery could prolong the creation process. For instance, E9 achieved the final design after three attempts, experimenting with different typeface granularities, including ``\textit{Bea}'', single ``\textit{e}'' and ``a'', and ``\textit{ea}''.
Participants noted that, apart from trying different parts of the typeface based on the selected imagery, it would also be challenging to find suitable imagery based on the chosen typeface."
}
% One possible way is employing a data-driven approach to establish a reasonable mapping between typeface and imagery using a large dataset.
% Aligning with established design practices will increase the likelihood of acceptance and appropriateness of the recommendations. 
% As it will have an ethical problem about innovation in the design, another way to implement this is based on structural similarity between the typeface and imagery.
% Prior research has explored retrieving visually pleasing image elements, such as icons and clipart, that share similar shapes with the typeface~\cite{zhang2017synthesizing,tendulkar2019trick, zhao2020iconate}. This approach can make the selection and generation process more targeted and reasonable. Users will have clearer expectations and anticipate the results more effectively.
% }

\subsubsection{Tradeoff between Imagery Diversity and Result Style Consistency}
\label{ssec: limitation1}
Fig.~\ref{fig:limitation} displays this limitation, where using the same imagery for \textit{``Hong Kong''} produces stylistically consistent result, while using different imagery leads to noticeable inconsistency.
E1, remarked, \textit{``These elements look good individually, but when combined, they appear discordant.''}
\blue{
This inconsistency arises from transferring imagery and style from image references to the generated result, resulting in various styles when using multiple references.
While adding a textual prompt is a partial solution to alleviate this issue, it lacks precise control.}
Incorporating multiple imageries within a single typeface is a common and significant format for semantic typographic logos. 
Thus, Achieving precise control over imagery diversity and result style consistency remains an important area for further investigation.
% Incorporating multiple concepts within a single typeface is a common and significant format for semantic typographic logos. 
% Thus, Achieving precise control over imagery diversity and result style consistency remains an important area for further investigation.

\subsubsection{Tradeoff between Typeface Complexity and Result Legibility}
\label{ssec: limitation2}
When the complexity of the typeface used in the creation rises (e.g., increasing the strokes or letters), the legibility of the generated result may decrease correspondingly.
As the right side of Fig.~\ref{fig:limitation} shows, using the same imagery during the creative process, the illustration of ``\raisebox{-.15\height}{\includegraphics[height=0.26cm]{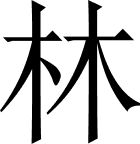}}'' is easily readable, while the presentation of ``\raisebox{-.15\height}{\includegraphics[height=0.26cm]{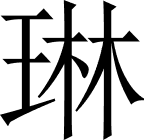}}'' appears more abstract.
% with the ``\raisebox{-.15\height}{\includegraphics[height=0.26cm]{figures/icons/characters-wang.png}}'' is nearly unrecognizable.
% P7 stated that ``the more complexity one looks pretty beautiful but it is hard to recognize and even be mislead to another word.''
\blue{
It suggests that the current generation ability of TypeDance can not perform well in cases with complex typeface, \textit{e.g.}, an entire Chinese word, or multiple letters.
The primary reason is that a complex typeface offers an initially rich structure for the generative model to stylize. To tackle this issue, TypeDance should find a viable solution to enhance control over the stylization progress during generation, achieving a balance between typeface complexity and legibility.}

% How to achieve strong legibility for high-complexity typefaces in semantic typographic logo design, that is, striking a balance between complexity and legibility, is also a topic the current system needs to explore.

%% file: Latex/7_Discussion.tex
\section{Discussion}
\label{sec:discussion}
% Based on comprehensive user and expert feedback, we have identified several opportunities for improving TypeDance in four main aspects: 1) \textit{domain knowledge}: incorporating more design principles, 2) \textit{generation performance}: improving style consistency among different imagery, 3) \textit{model robustness}: enhancing the model robustness towards complex typeface, and 4) \textit{system usability}: supporting selection recommendation.

\blue{
\subsection{Personalized Design: Intent-aware Collaboration with AI}
% text-authoring 
The rise of large language models has fueled a surge in text-driven creativity design~\cite{cao2023dataparticles, wang2022towards, yan2023xcreation}, enabling creators to collaborate with AI using natural language narratives.
% 好处是什么
While text-driven creation offers an intuitive means to manipulate the model in the backend without delving into complex parameters manually, expressing user intent concisely through textual prompts poses a challenge. Crafting a prompt becomes particularly daunting when describing an imagined visual design, given the myriad details such as layout, color, and shape that extend beyond textual representation.
PromptPaint~\cite{chung2023promptpaint} recognizes this challenge and approaches it by mixing a set of textual prompts to capture ambiguous concepts, like ``\textit{a bit less vivid color.}'' However, it remains constrained by offering a predefined set of prompts and fundamentally fails to resolve the issue of representing concrete visual concepts through prompts.

% image - example - inspiration
To ensure that a creator's intention aligns seamlessly with AI collaboration, it is crucial to mirror real design practices with accessible design materials.
The common design material used by creators includes explorable galleries~\cite{zhang2021method}, sketches~\cite{chung2022talebrush}, and even photographs capturing our perception of the world~\cite{zhang2020dataquilt, lupi2018observe}.
These visual design materials encompass both explicit intentions, such as prominent semantics, and implicit aesthetic factors. In logo design, there is a pronounced emphasis on identity, using images frequently to convey intentions. This aspect can not be ignored in AI collaboration, demanding a capability for AI to comprehend visual semantics.
It reveals that there is no universally superior material to encapsulate a creator's intent; it depends on the design task. This necessitates a hybrid and multimodal collaboration that can flexibly generalize to a wide array of requirements.
}

\blue{
\subsection{Incorporating Design Knowledge into Creativity Support Tools}
% We observed that integrating design knowledge involves two main aspects: \textit{generalizable design patterns} and \textit{simulatable design workflow}.
Instilling the generalizable design pattern into tools necessitates addressing \textit{technical} and \textit{interaction} challenges regarding how humans guide the model.
AI models are often not built for the special design task, posing challenges in generalizing to complex patterns.
For example, considerable research~\cite{zhang2017synthesizing,tanveer2023ds} has delved into blending techniques for specific typeface granularity. However, creativity support tools are user-oriented, with more intricate design requirements, calling for advanced techniques to accommodate all levels of typeface granularity.
Instead of retraining a model, significant research has explored to change or add the interaction with models for incorporating design knowledge, such as crowd-powered design parameterization~\cite{koyama2014crowd} and intervention through intermediate representations~\cite{yan2022flatmagic}.

The technical and interaction aspects of incorporating design knowledge externalize the idea of ``\textit{balancing automation and control,}'' which is often noted by existing human-AI design guidelines~\cite{amershi2019guidelines, apple2021human, google2019people}.
%  Apple. 2021. Human Interface Guidelines - Machine Learning. https://developer.apple.com/design/human-interface-guidelines/machinelearning/overview/introduction/. Accessed: 2021-09-9.
%  Google. 2019. People + AI Guidebook. https://pair.withgoogle.com/
% This balance leads to two implications.
The incorporation of design knowledge controlled by creators partially addresses the issue of AI copyright.
Current generative models have faced criticism for sampling examples from the training set.
In TypeDance, users contribute design materials through images, allowing for a personalized foundation instead of direct replication from a predefined dataset. This approach not only enhances creativity but also helps establish a stronger sense of ownership for creators.
Complete automation with a single model to achieve an end-to-end result overlooks the user's value. The allure of a creativity support tool, as opposed to relying solely on a model, lies in enabling creators to participate in crucial stages. This involvement includes customizing design materials, choosing which design knowledge to transfer to the generation process, and refining the final outcome.

% Therefore, in such a tool, the user plays the role of a creator, emphasizing the importance of their contribution throughout the creative process.
% Therefore, the role of user in such tool is creator
}

\blue{
\subsection{Mix-User Oriented Design Workflow}
With the aim of developing a tool with a ``low threshold'' for novices to steer the generation and a ``high ceiling'' for experts to achieve more advanced effects, TypeDance integrates a simulatable design workflow.
Creativity support tools are inherently designed to provide a comprehensive authoring experience, addressing both common and unique emphases from diverse users~\cite{zhou2023filtered, yan2022flatmagic, yan2023xcreation}.
As illustrated in Figure~\ref{fig:rating}, experts and designers share both similar and distinct preferences. Functions with shared preferences, like selection and generation, can be considered central to the workflow and warrant deeper investigation.
Notably, there are functions with varying preferences based on expertise. Experts tend to prioritize evaluation and refinement, whereas novices may view these as optional.
However, with less design background, novices find ideation helpful than experts.
Functions with differing preferences act as a  ``\textit{wide wall},'' accommodating optional user requirements. 
While not mandatory like central functions, omitting them compromises the overall integrity of the workflow.
}

%% file: Latex/8_Conclusion.tex
\section{Conclusion}
\label{sec:conclusion}

This study distills design knowledge from real-world examples, summarizes generalizable design patterns and simulatable design workflow, and explores the creation of semantic typographic logos by blending typeface and imagery while maintaining legibility.
We introduce TypeDance, an authoring tool based on a generative model that supports a personalized design workflow including ideation, selection, generation, evaluation, and iteration. With TypeDance, creators can flexibly choose typefaces at different levels of granularity and blend them with specific imagery using combinable design factors.
TypeDance also allows users to adjust the generated results along the typeface-imagery spectrum and offers post-editing for individual elements. Feedback from general users and experts validates the effectiveness of TypeDance and provides valuable insights for future opportunities.
We are excited to enhance the functionality of TypeDance for a comprehensive workflow and explore new techniques and interactions to enhance human creativity.